# ANA: Ant Nesting Algorithm for Optimizing Real-World Problems

**Deeam Najmadeen Hama Rashid** [1,*], **Tarik A. Rashid** [1,*] **and Seyedali Mirjalili** [2,3]

1 Department of Computer Science and Engineering, School of Science and Engineering, University of Kurdistan Hewler, 44001, Erbil, KRG, Iraq
2 Centre for Artificial Intelligence Research and Optimisation, Torrens University, Australia; ali.mirjalili@gmail.com
3 Yonsei Frontier Lab, Yonsei University, Seoul, Korea
* Correspondence: d.najmadeen@ukh.edu.krd (D.N.H.); tarik.ahmed@ukh.edu.krd (T.A.R.)

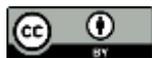

**Abstract:** In this paper, a novel swarm intelligent algorithm is proposed called ant nesting algorithm (ANA). The algorithm is inspired by Leptothorax ants and mimics the behavior of ants searching for positions to deposit grains while building a new nest. Although the algorithm is inspired by the swarming behavior of ants, it does not have any algorithmic similarity with the ant colony optimization (ACO) algorithm. It is worth mentioning that ANA is considered a continuous algorithm that updates the search agent position by adding the rate of change (e.g., step or velocity). ANA computes the rate of change differently as it uses previous, current solutions, fitness values during the optimization process to generate weights by utilizing the Pythagorean theorem. These weights drive the search agents during the exploration and exploitation phases. The ANA algorithm is benchmarked on 26 well-known test functions, and the results are verified by a comparative study with genetic algorithm (GA), particle swarm optimization (PSO), dragonfly algorithm (DA), five modified versions of PSO, whale optimization algorithm (WOA), salp swarm algorithm (SSA), and fitness dependent optimizer (FDO). ANA outperformances these prominent metaheuristic algorithms on several test cases and provides quite competitive results. Finally, the algorithm is employed for optimizing two well-known real-world engineering problems: antenna array design and frequency-modulated synthesis. The results on the engineering case studies demonstrate the proposed algorithm's capability in optimizing real-world problems.

## 1. Introduction

Both our professional and private life is a sequence of decisions and each decision involves selecting between at least two options (it will not be considered a decision otherwise), and the fact is we are always in search of finding the best option. Every question that needs a superlative answer is an optimization problem. Finding the shortest path to a destination, constructing the fastest car, hiring the best job applicant, prescribing the best medicine for a patient, making the best vaccine for a virus, finding the best way to discover diseases and viruses as early and reliable as possible, and finding the best strategy for overcoming financial and economic crises, are few examples of optimization problems.

Simple exhaustive search methods [1,2] are rarely sufficient for most real-world problems, and they lead to too slow or incomplete searches as the search space (the number of options) increases dramatically. That means, finding the best solution for a problem is not usually an easy task and requires a long time and sometimes an enormous amount of resources. Optimization can be defined as the art and science of making good decisions, and optimization algorithms are meant to solve optimization problems by trading in solution quality for runtime. Optimization algorithms provide us with a set of tools and techniques mainly from mathematics and computer science to select the best solution amongst the possible choices [3].

There are many ways in which optimization algorithms can be classified, the simplest way is to assort them as deterministic or stochastic [3]. Deterministic algorithms [3,4] such as linear programming, nonlinear programming, and mixed-integer nonlinear programming guarantee optimal or near-optimal solutions by adopting repeated design variables and functions firmly. They have a fast convergence rate and are simple and easy to implement and understand. Despite their efficiency [5], they are deterministic, i.e., for the same inputs, the same output is obtained consistently.

On the other hand, stochastic algorithms are more flexible and efficient than deterministic algorithms [3,6,7] as they are stochastic, i.e., they all have some level of randomness; for the same set of inputs, the same output is not always obtained. They are considered to be quite efficient in obtaining near-optimal solutions to all types of problems because they do not assume the underlying fitness landscape. The stochastic algorithms include heuristic and metaheuristic algorithms. Metaheuristic can be seen as a "*master strategy that guides and modifies other heuristics to produce solutions beyond those that are normally generated in a quest for local optimality*" [8]. Metaheuristics are considered to perform better than heuristics though the names are used interchangeably [3].

Time, cost, and resource limitations make searching every single solution for a problem to find the optimal one an impossible task. Therefore, researchers have started observing and studying the behavior of animals and natural phenomena to develop algorithms for solving optimization problems. They have developed algorithms based on swarm intelligence, biological systems, physical and chemical systems. These types of algorithms are called nature-inspired, and they contain a big set of novel problem-solving methodologies and approaches. They have been used to solve many real-world problems, and they comprise a large portion of stochastic algorithms.

From the time of advent, nature-inspired optimization algorithms have received great attention and are growing very rapidly. According to a research report [9], there are more than 200 nature-inspired algorithms presented in the literature. Despite the considerable number of algorithms and developments, there is always room for presenting a new algorithm, as long as the new proposed algorithm presents better or comparative performance to the previous ones as proven by the NFL theorem [10]. The theorem states that if any algorithm A outperforms another algorithm B in the search for an optimum of an objective, then algorithm B outperforms A over other objective functions. In other words, all the optimization algorithms give the same average performance when averaged over all functions. This proposes and motivates developing more and more algorithms for solving diverse and complex real-world optimization problems.

This paper proposes a new algorithm under the name ant nesting algorithm that is abbreviated to ANA. It is inspired by the swarming behavior of Leptothorax ants during nest construction. Ant colony optimization (ACO) algorithm is also a nature-inspired metaheuristic algorithm that mimics the behavior of ants. However, it is very different to our algorithm as it uses different ant stigmergy and behavior.

The major contribution of this work is a proposal of a new swarm intelligent algorithm for optimizing single-objective problems that has a good level of exploration and exploitation. It is noted that single-objective optimization problems are those problems that require a solution for a single criterion or metric of the problem.

The main contributions of this work are outlined as follows:

(1) Proposing a novel metaheuristic algorithm for solving SOPs.
(2) Integrating Pythagorean theorem into the ant nesting model for generating convenient weights that assist the algorithm in both exploration and exploitation phases.
(3) Utilizing a quite different approach from PSO for updating search agent positions and testing the algorithm on several optimization benchmark functions and comparing it to the most well-known and outstanding metaheuristic algorithms like a genetic algorithm (GA), particle swarm optimization (PSO), five modified versions of PSO, dragonfly algorithm (DA), whale optimization algorithm (WOA), salp swarm algorithm (SSA), and fitness dependent optimization (FDO).

(4)  Applying ANA algorithm for optimizing two real-world engineering problems that are antenna array design and frequency-modulated synthesis.

The rest of the paper contains a brief history of the most prominent swarm algorithms in the literature, the inspiration of the algorithm with its modelling, testing and evaluation of the algorithm, and the conclusion and recommendation of a few future works.

## 2. Nature-Inspired Metaheuristic Algorithms in Literature

Metaheuristic as a scientific method for solving problems is a quite new phenomenon in comparison to its ubiquitous nature. Although it is difficult to pinpoint the first use of metaheuristics, the mathematician Alan Turing is known to be the first to have used heuristic algorithms for breaking the Enigma cyphers at Bletchley Park during World War II [3].

To date, over 200 nature-inspired metaheuristic algorithms for optimization exist in the literature. This section presents the most well-known algorithms in the literature. Figure 1 highlights the year of development of each algorithm mentioned.

A genetic algorithm was developed by Holland and his colleagues at the University of Michigan in the 1960s and 1970s [11]. It is based on the abstraction of Darwinian evolution and the natural selection of biological systems. GA is proven to be extremely successful in solving a wide range of optimization problems, and hundreds and thousands of books and research papers have been published about it. In addition to that, some studies show its usage in solving combinatorial optimization problems like scheduling, planning, and data mining in companies [12].

Ant colony optimization was developed by Dorigo in 1992, and it is inspired by the foraging behavior of social ants; how ants find the shortest path to a destination [13]. ACO has been applied for solving various problems such as scheduling [14], vehicle routing [15], assignment [16], and set [17].

Particle swarm optimization was developed by Kennedy and Eberhart in 1995, and it is inspired by the movement behavior of bird flocks and fish schools [18]. The algorithm's exploitation and exploration abilities are considered to be quite efficient, and it has been used for solving a very large number of real-world optimization problems [19].

The artificial bee colony algorithm was developed by Karaboga in 2005, and it is inspired by the intelligent foraging behavior of honeybee swarms [20]. ABC has high exploration ability and comparatively low exploitation ability. Although the performance of the algorithm depends on applications, ABC has been proven to be more efficient than GA and PSO in solving certain problems [21–23]. Examples of applications are in solving the set covering problem (SCP) [24] and optimum reservoir release [25].

Firefly algorithm was developed by Yang in 2008, and it is inspired by the behavior of the flashing pattern of fireflies [26]. It has been used for solving a variety of optimization problems in computer science and engineering, and it has been proven to outperform some other metaheuristic algorithms. However, despite the application and efficiency of FA, it has been criticized as differing from PSO only in a negligible way [27–29].

The cuckoo search was proposed by Yang and Deb in 2009, and it is inspired by the brood parasitism of cuckoo species [30]. CS uses a switching parameter to balance between local and global random walks. It outperforms PSO and GA algorithms and is used for solving several real-world problems like power network planning [31], series system [32], and engineering design optimization problems [33].

Bat algorithm was developed by Yang in 2010, and it is based on the echolocation behavior of microbats [34]. BA has been applied in several areas like image processing [35] and scheduling [36].

Grey wolf optimizer was proposed by Mirajili et al. in 2014 and is inspired by the searching and hunting behavior of grey wolves. The algorithm has been implemented in three steps, which are searching, encircling, and attacking prey. GWO has been proven to be very efficient and outperforms several well-known algorithms like gravitational search algorithm (GSA), PSO, differential evolution (DE), evolutionary programming (EP), and evolution strategy (ES) on several benchmark test functions. It has been applied for solving classical engineering design problems and optical engineering real-world problems [37].

Dragonfly algorithm was developed by Mirajili in 2015 and is inspired by the static and dynamic swarming behaviors of dragonflies. For simulating the swarming behavior, the five swarming principles of insects have been utilized, which are alignment, cohesion, separation, attraction to a food source, and distraction from the enemy. DA is available to be used for solving single-objective, discrete, and multi-objective optimization problems [38].

Ant lion optimizer was proposed by Mirjalili in 2015 and is inspired by the hunting mechanism of antlions in nature. Five main steps of hunting prey like the random walk of ants, building traps, entrapment of ants in traps, catching prey, and re-building traps are implemented. ALO has been used for solving three-bar truss design, cantilever beam design, and gear train design and optimizing the shapes of two ship propellers [39].

The whale optimization algorithm (WOA) was developed by Mirjalili and Lewis in 2016 [40]. WOA mimics the hunting mechanism of humpback whales and has three phases, which are encircling prey, bubble-net attacking method, and search for prey. It has been applied for solving several optimization problems and provided outstanding results like economic dispatch problem [41], breast cancer diagnosis [42], global MPP tracking of a photovoltaic system [43], and a handful number of other significant problems [44–46].

Salp swarm algorithm was developed by Mirjalili et al. in 2017. It is inspired by the behavior of salp swarms when navigating and foraging in oceans. SSA has a single decreasing parameter to make the balance between diversification and intensification. It can be used for solving both single-objective and multi-objective optimization problems. It is applied for solving several challenging engineering designs [47].

Donkey and smuggler optimization algorithm was proposed by Shamsaldin et al. in 2019. DSO mimics the searching behavior of donkeys while transporting goods, and it consists of two modes that are the donkeys and the smuggler modes. In the smuggler mode, all the possible solutions are discovered and the best one is identified. While in the donkey's mode, a couple of donkeys' behaviors are utilized for finding the optimal option among the possible solutions. The algorithm has been applied to solve three well-known real-world optimization problems that are namely traveling salesman problem, packet routing, and ambulance routing, and significant results were produced by the algorithm [48].

Fitness dependent optimizer was developed by Abdullah and Rashid in 2019 [49]. It is a PSO-based algorithm and is inspired by the foraging behavior of honeybees whilst selecting a hive. It has been improved and applied by Muhammed et al. in 2020 to develop a model for pedestrian evacuation [50].

In addition to the standard versions of the nature-inspired metaheuristic algorithms, there are many modified, enhanced versions of them like GPSO, EPSO, LNPSO, VC-PSO, and SO-PSO that are modified versions of the PSO algorithm [51], and DCSO which is a modified version of CWO algorithm [52] including the hybridized algorithms like WOAGWO [53], that composes the behavior of more than one algorithm.

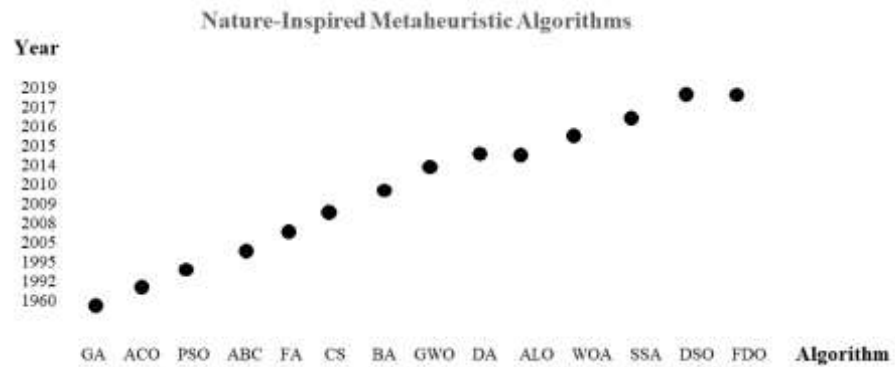

**Figure 1.** Some of the most prominent nature-inspired metaheuristic algorithms in literature.

## 3. Ant Swarming

Ants are remarkable tiny insects with an enormous population and over 10,000 species all over the world, and they have always been in the position of wonder and fascination to human beings. A considerable number of books and research papers have been published about all the aspects of their life, ranging from pure literature [54,55] to detailed scientific accounts [56,57]. What is striking the most is their fascinating high social organization in comparison to their limited individual capabilities. They have been an inspiration to researchers and scientists for years. This section shortly presents Leptothorax ant behavior while building a new nest.

Leptothorax ant colonies prefer horizontal crevices and build nests within flat crevices in rocks that provide the roof and floor of their dwelling place to defend themselves against physical factors and biological enemies. They wall themselves into their chosen crevice by encircling themselves with a border of debris or grains, which are particles of earth or small fragments of stones. The circumference of the wall is appropriate to comfortably house their population which consists of a single queen, broods, and up to 500 workers. Worker ants are responsible for building the nest, and each is 2 to 3 mm long [58,59]. Figure 2 demonstrates ant nesting.

Leptothorax worker ants do not build a roof or floor for their nest. They build their nest by only constructing a wall around their queen and broods that consist of eggs, larvae, and pupae. The cluster of worker ants around the brood cluster serves as a mechanical template for determining where the nest wall should be built [58,59].

The nest construction starts with the worker ants in the colony departing from their cluster, collecting a single grain of building material, and dropping it within a distance from the cluster randomly. Then, the ants lean towards the area with the most dropped stone and start bulldozing stones into other stones from that area. The process of selecting an area to start bulldozing is very important for the consolidation of the wall. The bulldozing process continues till a well tightly packed and densely consolidated wall is formed around the queen ant in the center [58,59].

What inspired in developing this algorithm is the worker ants' individual decision to drop grains at a fixed distance from the center of the nest with stigmergic interaction of deposing where others have already been deposited. The wall originates from a combination of each worker ant's decision. In other words, when constructing a new nest, worker ants select an area around the queen, the area with the most grain, and start bulldozing from that area. A decision is made when the majority of the ants are bulldozing at a potential area [58,59].

The Leptothorax worker ants' building behavior that inspired the algorithm is summarized as the following:

- Worker ants are responsible for building new nests by collecting building materials, transporting them into the nest site, and releasing them in an area around the queen ant [58,59].
- Worker ants make a random walk within the nest until they face their nestmates or stationary building materials to deposit; the latter is the major cue for the deposition of another building material [58].
- Each worker ant makes an independent decision about which direction to take around the queen ant for depositing [59].
- Worker ants lean towards the area with the most dropped building material to deposit [58].
- Each worker ant selects an area around the queen ant to start the bulldozing process. A decision is made when all the worker ants in the colony are bulldozing at a potential area, i.e., deposit grain in that area [58].

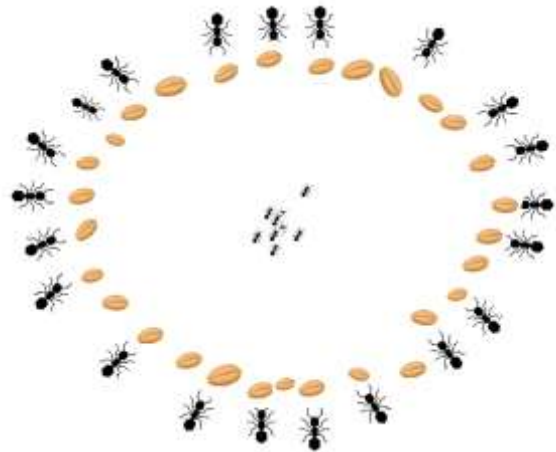

**Figure 2.** Ant nesting.

## 4. Ant Nesting Algorithm

In this section, the modelling process including the entities, mathematical representations, working mechanism, and analysis of the algorithm is provided.

### 4.1. Entities

Algorithmically modelling the entities of the Leptothorax ant behavior while building a new nest, the worker ants represent artificial search agents; each position around the queen ant that a worker ant exploits to drop grain, represents a potential solution exploited by an artificial search agent, and the best position to deposit among all the possible positions exploited by all the worker ants represents the global optimum solution. Accordingly, the deposition position is determined through the worker ants' position. The deposition position's specification, such as its influence on the consolidation of the wall and its closeness to other stones, can be considered as fitness functions of the algorithm. Each worker ant's decision factor to deposit grain at a specific position is represented by deposition weight ($dw$) in the algorithm. $dw$ is a random weight of worker ant to deposit grain at a specific position using solution and fitness information of the previous and current deposition positions, it is discussed further in the next section. Table 1 summarizes the main elements of the algorithm.

The stationary stones and nestmates that are encountered by the worker ants while performing random walks within the nest to find new better positions to drop grain are modelled using the worker ant's previous deposition position $X_{t,iprevious}$. That is to say that the current worker ant's previous deposition position represents the stationary stone and/or nestmate the current worker ant faces in the algorithm.

**Table 1.** entities in the ANA algorithm.

| Nature | Algorithm |
|---|---|
| Worker ant | Search agent |
| Deposition position | Potential solution |
| Deposition position specification | Fitness function |
| Worker ant's decision factor | Deposition weight |
| Fittest deposition position | Optimum solution |
| Stationary stone and/or nestmate | Previous deposition position |

*4.2. Mathematical Modelling*

The algorithm mimics what a swarm of worker ants is doing during nesting. The main part of the algorithm is taken from the process of worker ants searching for a suitable position among many potential positions to deposit grain. Every worker ant that searches for deposition positions represents a potential solution in this algorithm; furthermore, selecting the best deposition position among several good deposition positions is considered as converging to optimality.

It is noteworthy that in nature, worker ants collect grains, transport them to the nest site, and search for deposition positions continuously as a cycle till a well tightly wall is formed around the queen ant. However, only a single cycle of dropping grain is modelled in the algorithm, i.e., the algorithm only mimics the worker ants search for the deposition position of a single grain during nesting not continuously searching for deposition positions for each grain collected by each worker ant. In addition to that, the first building workers who use the brood cluster as a mechanical template for determining where the nest wall should be built are ignored in the algorithm. Rather, the grain dropped after the first depositions are considered in modelling the algorithm.

The algorithm begins by randomly initializing the artificial worker ant population in the search space $X_i$ ($i$ = 1, 2, 3, …, $n$); each worker ant position represents a newly discovered deposition position (solution). Worker ants try to find better deposition positions by randomly searching more positions. Each time a better deposition position is found, the newly discovered solution becomes the optimum solution. However, if the new solution is not better than the current, it will then continue to the current solution, which is the best solution that has been discovered to that point.

In nature, worker ants search for deposition positions randomly. In this algorithm, artificial worker ants search the landscape randomly using a deposition weight mechanism. Accordingly, every time an artificial worker ant obtains a new deposition position $X_{t+1,i}$ ($t$ = 1, 2, 3, …, $m$) ($i$ = 1, 2, 3, …, $n$) by adding deposition position rate of change that is denoted by $\Delta X_{t+1,i}$, to their current deposition position $X_{t,i}$. The deposition position of the artificial worker ant is updated with the following expression:

$$X_{t+1,i} = X_{t,i} + \Delta X_{t+1,i} \tag{1}$$

where, $i$ represents the current worker ant, $t$ represents the current iteration, $X$ represents the artificial worker ant's deposition position, and $\Delta X_{t+1,i}$ represents the deposition position's rate of change. Table 2 summarizes all the mathematical notations used in the algorithm.

The deposition position's rate of change $\Delta X_{t+1,i}$ is dependent on deposition weight $dw$ and the difference between the local best-known worker ant $X_{t,ibest}$ and deposition position of the current worker ant $X_{t,i}$; the latter is the mathematical modelling of the behavior of leaning towards the most dropped building material. Thus, each worker ant tends to improve its deposition position (potential solution) by moving towards the best-known worker ant (the best potential solution discovered so far). Thereby, the $\Delta X_{t+1,i}$ is calculated as the following:

$$\Delta X_{t+1,i} = dw \times (X_{t,ibest} - X_{t,i}) \tag{2}$$

The following rule is followed for calculating $\Delta X_{t+1,i}$ when: the current worker ant is the local best-known ant

$$\Delta X_{t+1,i} = r \times X_{t,i} \qquad (3)$$

the current deposition position is equal to the previous deposition position

$$\Delta X_{t+1,i} = r \times (X_{t,ibest} - X_{t,i}) \qquad (4)$$

**Table 2.** ANA's mathematical notations.

| Notation | Description |
|---|---|
| $t$ | Current iteration |
| $i$ | Current worker ant |
| $m$ | Iteration number |
| $n$ | Worker ant number |
| $X_{t,i}$ | Worker ant's current deposition position |
| $X_{t,ibest}$ | Worker ant's local best deposition position |
| $X_{t,iprevious}$ | Worker ant's previous deposition position |
| $X_{t,i}$ fitness | Worker ant's current deposition position's fitness |
| $X_{t,ibest}$ fitness | Worker ant's local best deposition position fitness |
| $X_{t,iprevious}$ fitness | Worker ant's previous deposition position fitness |
| $X_{t+1,i}$ | Worker ant's new deposition position |
| $\Delta X_{t+1,i}$ | Worker ant's deposition position's rate of change |
| $T$ | Worker ant's current deposition tendency rate |
| $T_{previous}$ | Worker ant's previous deposition tendency rate |
| $dw$ | Deposition weight |
| $r$ | Random number in the [−1, 1] range |

Deposition weight ($dw$) is the mathematical representation of the random walk performed by the worker ant, and it depends on the artificial worker ant's previous ($T_{previous}$) and current ($T$) tendency rate to deposit grain at a specific position. $T$ and $T_{previous}$ are computed as the slope sides in the Pythagorean theorem of the difference between the worker ant's current and previous deposition positions to the best deposition position discovered so far with their fitness difference as the other sides. Figure 3a,b explicitly demonstrates how the $T$ and $T_{previous}$ are gained for a single ant respectively. Thus, $dw$ for minimization problems can be calculated as the following:

$$dw = r \times \left( \frac{T}{T_{previous}} \right) \qquad (5)$$

where, $r$ is a random number in [−1, 1] range, works as deposition factor for controlling the $dw$. There are different mechanisms for generating random numbers. Levy flight has been selected because it provides more stable movements due to its good distribution curve [26].

The worker ant's tendency rate of deposition ($T$) is calculated as the following:

$$T = \sqrt{(X_{t,ibest} - X_{t,i})^2 - (X_{t,ibest} fitness - X_{t,i} fitness)^2} \qquad (6)$$

The worker ant's previous tendency rate of deposition ($T_{previous}$) is calculated as the following:

$$T_{previous} = \sqrt{(X_{t,ibest} - X_{t,iprevious})^2 - (X_{t,ibest} fitness - X_{t,iprevious} fitness)^2} \qquad (7)$$

### 4.3. Working Mechanism

The algorithm starts by initializing random deposition positions $X_{t,i}$ ($t$ = 1, 2, 3, …, $m$) ($i$ = 1, 2, 3, …, $n$) for each artificial worker ant using the lower and upper boundaries. Initially, the previous deposition position $X_{t,iprevious}$ of each artificial worker ant is assigned to $X_{t,i}$ as it is the first generation. Then for each iteration, the global best deposition position $X_{t,ibest}$ is selected, a random number $r$ in the range [−1, 1] will be generated, and for each artificial worker ant, $X_{t,i}$ is compared to $X_{t,ibest}$. If the current worker ant deposition position is the global best solution discovered so far, i.e., $X_{t,i}$ is equal to $X_{t,ibest}$, calculate $\Delta X_{t+1,i}$ using Equation (3), and if the previous deposition position is equal to current deposition position, calculate $\Delta X_{t+1,i}$ using Equation (4). Otherwise, calculate $T$, $T_{previous}$, $dw$, and $\Delta X_{t+1,i}$ using Equations (6), (7), (5) and (2) respectively.

After that, a new solution $X_{t+1,i}$ is obtained through Equation (1). Each time the artificial worker ant finds a new solution, it checks whether the new solution is better than the current solution using the fitness function. If the new solution is fitter, then it is accepted and the old solution is saved to $X_{t,iprevious}$. However, if the new solution is not fitter, then the algorithm maintains the current solution until the next iteration. For elucidating the working mechanism of the ANA algorithm more, both pseudocode and flowchart are developed. Figure 4 shows the pseudocode of the ANA algorithm, and Figure 5 presents the flowchart.

When implementing the ANA algorithm for maximization problems, two minor changes are needed. First, Equation (5) must be replaced by Equation (8), as Equation (8) is simply an inverse version of Equation (5).

$$dw = r \times \left( \frac{T_{previous}}{T} \right) \qquad (8)$$

Second, the condition for selecting a better (fitter) solution should be changed.

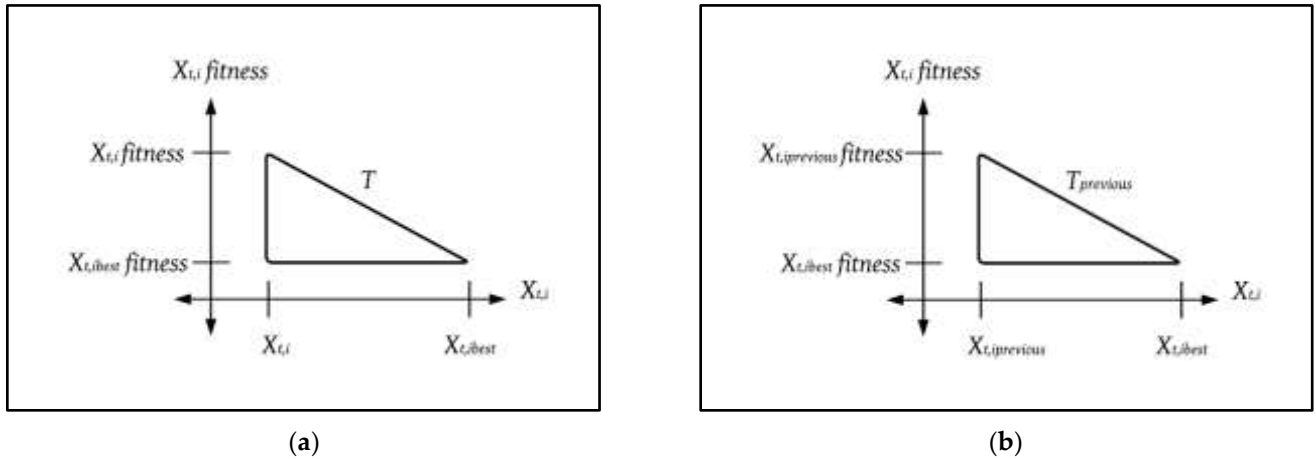

(**a**)                                                    (**b**)

**Figure 3.** $T$ and $T_{previous}$ computation: (**a**) $T$ is computed as the slope side of the difference between the current and best worker ants' deposition positions as one side and their fitness difference as the other side; (**b**) $T_{previous}$ is computed as the slope side of the difference between the current worker ant's previous and the best deposition positions as one side and their fitness difference as the other side.

Initialize worker ant population $X_i$ ($i$ = 1, 2, 3, …, $n$)

Initialize worker ant previous position $X_{iprevious}$

**while** iteration ($t$) limit not reached

   **for** each artificial worker ant $X_{t,i}$

      find best artificial worker ant $X_{t,ibest}$

      generate random walk $r$ in [-1, 1] range

      **if** ($X_{t,i}$ == $X_{t,ibest}$)

         calculate $\Delta X_{t+1,i}$ using Equation (3)

      **else if** ($X_{t,i}$ = $X_{t,iprevious}$)

         calculate $\Delta X_{t+1,i}$ using Equation (4)

      **else**

         calculate $T$ using Equation (6)

         calculate $T_{previous}$ using Equation (7)

         calculate $dw$ using Equation (5)     // for minimization

         calculate $\Delta X_{t+1,i}$ using Equation (2)

      **end if**

         calculate $X_{t+1,i}$ using Equation (1)

      **if** ($X_{t+1,i}$ *fitness* < $X_{t,i}$ *fitness*)    // for minimization

         move accepted and $X_{t,i}$ assigned to $X_{t,iprevious}$

      **else**

         maintain current position

      **end if**

   end for

end while

**Figure 4.** Pseudocode of ANA for a minimization problem without the loss of generality.

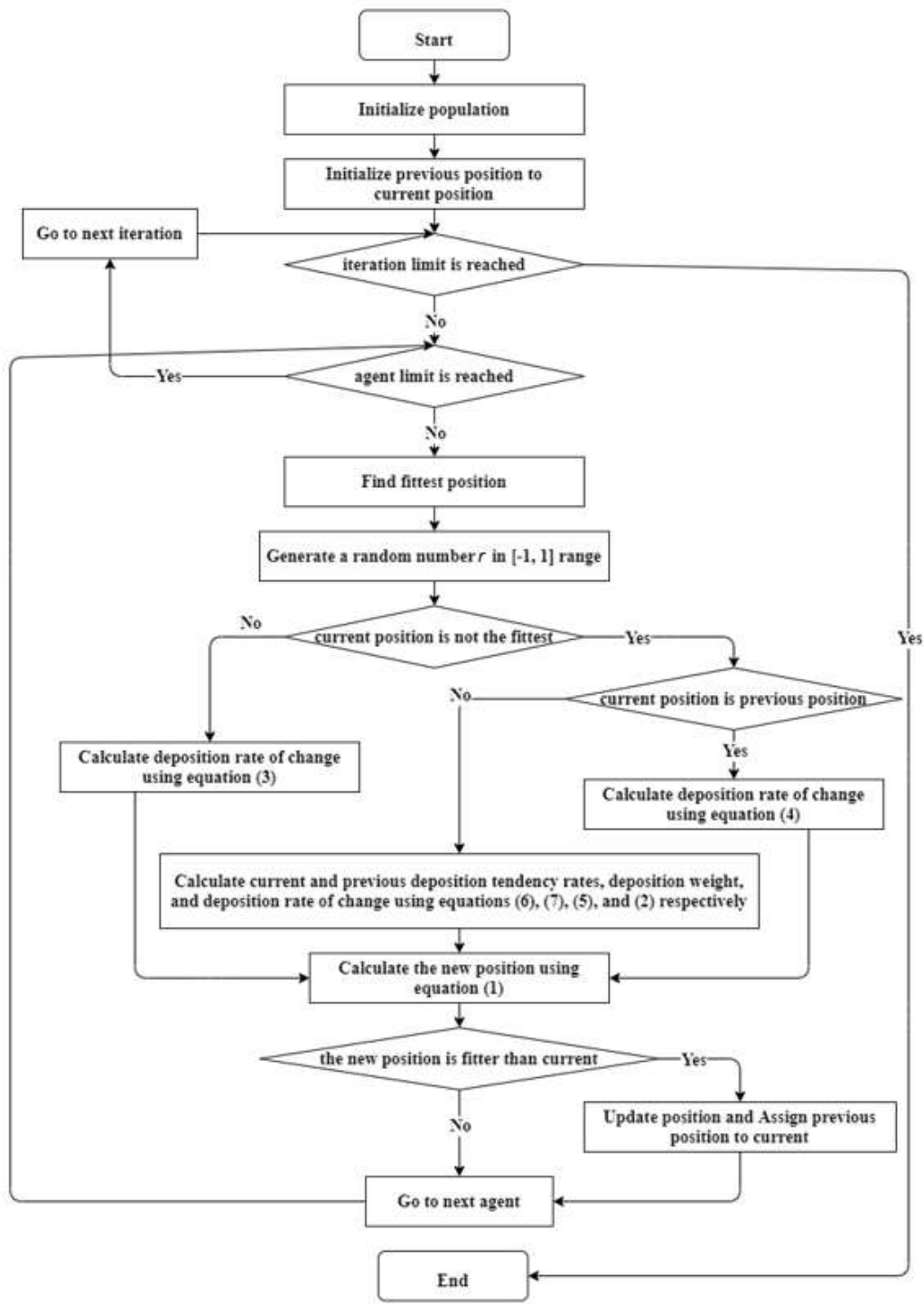

**Figure 5.** Flowchart of ANA.

## 5. Testing and Evaluation

For measuring and evaluating the performance and feasibility of an algorithm, several techniques exist in the literature including test functions and real-world applications. This chapter presents the test mechanisms used with the results and analysis of the results.

*5.1. Standard Benchmark Functions*

A considerable number of common standard benchmarks or test functions for testing reliability, efficiency, and validation of optimization algorithms exist. However, the effectiveness of an algorithm against others cannot be measured by the problems, which it solves if the set of problems selected are too specific and do not have varied properties. Therefore, a set of 16 common standard benchmark functions with diverse characteristics are selected for testing the performance of the algorithm. The sets consist of unimodal, multimodal, and composite test functions. Tables A1–A3 in Appendix A demonstrate the selected standard benchmark functions for benchmarking performance.

The unimodal test functions have only a single optimum. They are used for testing exploitation ability. They allow focusing more on the convergence rate of the algorithm other than the final results. Table A1 in Appendix A presents the six unimodal test functions selected for testing the ANA algorithm that are namely F1, F2, F3, F4, F5, and F7. F6 is not selected as a benchmark function for testing the algorithm but is mentioned in the table because it is utilized in the composite functions [38].

The multimodal test functions have more than one optimum, and the number of local optima usually increases exponentially with the number of problem dimensions. They are used for testing exploration ability which can make the algorithm avoid local optima/s. Table A2 in Appendix A presents the four multimodal test functions selected for testing the ANA algorithm that are namely F9, F10, F11, and F12. F8 is not selected as a benchmark function for testing the ANA algorithm; however, it is mentioned in the table because it is utilized in the composite functions [38].

The composite test functions as the name implies, are the combined, shifted, rotated, and biased versions of the other test functions. They provide a lot of varied shapes with several local optima. They allow measuring the exploitation and exploration balance of the algorithm. Table A3 in Appendix A presents the six composite test functions selected for testing the algorithm [38]. For verification and analysis purposes, our proposed ANA algorithm is compared to two sets of competitive algorithms with different parameter settings for each set.

### 5.1.1. DA, PSO, and GA

In the first set, the standard GA, PSO, and DA algorithms are selected as reference algorithms for comparison to ANA on the 16 selected standard benchmark functions. GA and PSO are among the earliest well-known and efficient algorithms in literature [60], and DA is a recently developed, promising algorithm with a high number of successful applications [61]. The reference [38] contains the test results of DA, PSO, and GA algorithms on the 16 selected standard benchmark functions including the parameter settings in detail.

Regarding the common parameter sets in all the cases, the population size is set to 30, and the dimension of the benchmark functions is equal to 10. The maximum number of iterations is set as the stopping criteria, equal to 500. The algorithm is tested 30 times and the average and standard deviation are calculated. Table 3 presents the test results of ANA, DA, PSO, and GA algorithms on the 16 standard benchmark functions [38].

Each test function of the algorithm from the standard benchmark functions is minimized towards 0.0 as shown in Table 3. Comparing the test results of ANA to the other algorithms presented in the table; ANA outperformed the most famous algorithms: DA, PSO, and GA on seven test cases are namely F4, F10, F13–16, and F18, and only on F7 the other algorithms provided better results than ANA. However, the result of F7 was not poor but only not better than the other algorithms. On the rest of the benchmark functions, the algorithm provided comparative results to the others.

According to the results of Table 3, it can be noted that F13 to F18 are composite test functions and are suitable for measuring the local minima avoidance of an algorithm. ANA algorithm outperformed all the DA, PSO, and GA algorithms on all these test functions except F17 that came in the third position with the outperformance of the DA algorithm. From that, it can be concluded that ANA is quite effective in avoiding local minima and balancing exploitation and exploration levels.

**Table 3.** ANA, DA, PSO, and GA test results on the standard benchmark functions [38].

| Test Function | ANA | | DA [38] | | PSO [38] | | GA [38] | |
|---|---|---|---|---|---|---|---|---|
| | Mean | Standard Deviation | Mean | Standard Deviation | Mean | Standard Deviation | Mean | Standard Deviation |
| F1 | 0.016299162 | 0.062457134 | **2.85 × 10⁻¹⁸** | $7.16 \times 10^{-18}$ | $4.20 \times 10^{-18}$ | $1.31 \times 10^{-17}$ | 748.5972 | 324.9262 |
| F2 | $2.36 \times 10^{-5}$ | $2.15 \times 10^{-5}$ | **1.49 × 10⁻⁵** | $3.76 \times 10^{-5}$ | 0.003154 | 0.009811 | 5.971358 | 1.533102 |
| F3 | 1.239172335 | 1.035406371 | **1.29×10⁻⁶** | $2.10 \times 10^{-6}$ | 0.001891 | 0.003311 | 1949.003 | 994.2733 |
| F4 | **2.37×10⁻⁶** | $8.86 \times 10^{-16}$ | 0.000988 | 0.002776 | 0.001748 | 0.002515 | 21.16304 | 2.605406 |
| F5 | 14.95306041 | 30.63521072 | **7.600558** | 6.786473 | 63.45331 | 80.12726 | 133307.1 | 85,007.62 |
| F7 | 0.770696384 | 0.366042632 | 0.010293 | 0.004691 | **0.005973** | 0.003583 | 0.166872 | 0.072571 |
| F9 | 25.46221873 | 4.313900987 | 16.01883 | 9.479113 | **10.44724** | 7.879807 | 25.51886 | 6.66936 |
| F10 | **5.54 × 10⁻¹⁵** | $2.37 \times 10^{-15}$ | 0.23103 | 0.487053 | 0.280137 | 0.601817 | 9.498785 | 1.271393 |
| F11 | 0.411189712 | 0.073239096 | 0.193354 | 0.073495 | **0.083463** | 0.035067 | 7.719959 | 3.62607 |
| F12 | 3.219956841 | 2.52657309 | 0.031101 | 0.098349 | **8.57 × 10⁻¹¹** | $2.71 \times 10^{-10}$ | 1858.502 | 5820.215 |
| F13 | **1.76 × 10⁻²³** | $9.47 \times 10^{-23}$ | 103.742 | 91.24364 | 150 | 135.4006 | 130.0991 | 21.32037 |
| F14 | **4.26 × 10⁻¹⁴** | $1.54 \times 10^{-13}$ | 193.0171 | 80.6332 | 188.1951 | 157.2834 | 116.0554 | 19.19351 |
| F15 | **4.89 × 10⁻⁶** | $3.31 \times 10^{-6}$ | 458.2962 | 165.3724 | 263.0948 | 187.1352 | 383.9184 | 36.60532 |
| F16 | **23.76092355** | 0.048390796 | 596.6629 | 171.0631 | 466.5429 | 180.9493 | 503.0485 | 35.79406 |
| F17 | 223.5622125 | 0.008813889 | 229.9515 | 184.6095 | 136.1759 | 160.0187 | **118.438** | 51.00183 |
| F18 | **31.51015225** | 0.020777872 | 679.588 | 199.4014 | 741.6341 | 206.7296 | 544.1018 | 13.30161 |

\* The bold values of the mean indicate the best solution has been obtained by the algorithm in comparison to the others.

5.1.2. PSO, GPSO, EPSO, LNPSO, VC-PSO, and SO-PSO

In the second set, the standard PSO, and several modified versions of the PSO algorithm that are GPSO, EPSO, LNPSO, VC-PSO, and SO-PSO are selected as reference algorithms for comparing to ANA on six selected standard benchmark functions that are namely F1, F5, F7, F9, F10, and F11. This work [51] contains the detail of modification of the GPSO, EPSO, LNPSO, VC-PSO, and SO-PSO algorithms, parameter settings, and test results of the algorithms on the selected functions.

Regarding the common parameter sets in all the cases, the population size is set to 30, and the dimension of the benchmark functions is equal to 20. The maximum number of iterations is set as the stopping criteria, equal to 10,000. It is worth mentioning that the functions are used without shift and with the same range of the first set except for F1 the range is reduced to [−5.12, 5.12]. The algorithm is tested 100 times and the average and standard deviation are calculated. Table 4 presents the test results of ANA, PSO, GPSO, EPSO, LNPSO, VC-PSO, and SO-PSO algorithms on the six selected standard benchmark functions [51].

Each test function of the algorithm from the standard benchmark functions is outstandingly minimized towards 0.0 as shown in Table 4. Comparing the test results of ANA to the other algorithms presented in the table; ANA outperformed these algorithms: PSO, GPSO, EPSO, LNPSO, VC-PSO, and SO-PSO on two test cases that are namely F1 and F9. On the rest of the benchmark functions, the ANA algorithm provided quite comparative results to the others.

**Table 4.** ANA, PSO, GPSO, EPSO, LNPSO, VC-PSO, and SO-PSO test results on the standard benchmark functions [51].

| Test Function | | ANA (this work) | PSO [51] | GPSO [51] | EPSO [51] | LNPSO [51] | VC-PSO [51] | SO-PSO [51] |
|---|---|---|---|---|---|---|---|---|
| F1 | **Mean** | 0 | $1.17 \times 10^{-45}$ | $1.11 \times 10^{-45}$ | $1.17 \times 10^{-45}$ | $1.11 \times 10^{-45}$ | $1.17 \times 10^{-108}$ | $1.51 \times 10^{-108}$ |
| | Standard deviation | 0 | $5.22 \times 10^{-46}$ | $4.76 \times 10^{-46}$ | $5.22 \times 10^{-46}$ | $4.76 \times 10^{-46}$ | $4.36 \times 10^{-108}$ | $4.46 \times 10^{-108}$ |
| F5 | Mean | 14.7067849 | 22.19173 | 9.99284 | 8.995165 | **4.405738** | 6.30326 | 6.81079 |
| | Standard deviation | 0.17792416 | $1.62 \times 10^{4}$ | 3.16891 | 3.959364 | 4.121244 | 3.99428 | 3.76973 |
| F7 | Mean | 0.84120263 | 8.681602 | 0.63602 | **0.380297** | 0.537461 | 0.410042 | 0.806175 |
| | Standard deviation | 0.55592343 | 9.001534 | 0.29658 | 0.281234 | 0.285361 | 0.294763 | 0.868211 |
| F9 | Mean | **$1.12 \times 10^{-7}$** | 22.33916 | 9.75054 | 12.17397 | 23.50713 | 9.99929 | 8.95459 |
| | Standard deviation | $3.32 \times 10^{-7}$ | 15.93204 | 5.43379 | 9.274301 | 15.30457 | 4.08386 | 2.65114 |
| F10 | Mean | $5.42 \times 10^{-15}$ | $3.48 \times 10^{-18}$ | $3.14 \times 10^{-18}$ | $3.37 \times 10^{-18}$ | $3.37 \times 10^{-18}$ | $5.47 \times 10^{-19}$ | **$4.59 \times 10^{-19}$** |
| | Standard deviation | $1.74 \times 10^{-15}$ | $8.36 \times 10^{-19}$ | $8.60 \times 10^{-19}$ | $8.60 \times 10^{-19}$ | $8.60 \times 10^{-19}$ | $1.78 \times 10^{-18}$ | $1.54 \times 10^{-18}$ |
| F11 | Mean | 0.92650251 | 0.031646 | 0.00475 | 0.011611 | 0.011009 | **0.00147** | 0.001847 |
| | Standard deviation | 0.02222257 | 0.025322 | 0.01267 | 0.019728 | 0.019186 | 0.00469 | 0.004855 |

* The bold values of the mean indicate the best solution has been obtained by the algorithm in comparison to the others.

### 5.2. CEC-C06 2019 Benchmark Functions

In addition to the standard benchmark functions, a set of 10 modern CEC benchmark functions are used as an extra evaluation of the ANA algorithm, and the results are compared to three other remarkable metaheuristic algorithms that are DA, WOA, and SSA. These test functions are developed for benchmarking single-objective optimization problems, and they are known as "the 100-digit challenge" that is intended to be used in annual optimization competitions [62]. Table A4 in Appendix A presents the CEC-06 2019 test functions used for benchmarking the ANA algorithm [62].

All the test functions from CEC-06 2019 are scalable, and functions CEC01 to CEC03 are not shifted or rotated but functions CEC04 to CEC10 are. The default test function parameters that were set by the CEC benchmark developer is used for performing the test of ANA. As can be seen in Table A4 in Appendix A, function CEC01 is set as a 9-dimensional minimization problem in [−8192, 8192] boundary range, function CEC02 is set as 16 dimensional in [−16,384, 16,384] range, function CEC03 is set as 18 dimensional in [−4, 4] range, while the rest of the functions from CEC04 to CEC10 are set as a 10-dimensional minimization problem in [−100, 100] boundary range. All the CEC functions' global optimum were unified towards 1.0 for providing more convenience.

The test results of the ANA algorithm are compared to the test results of three other modern optimization algorithms that are DA, WOA, and SSA which are taken from Abdullah and Rashid [49]. Regarding common parameter settings, the same is used as the ones that have been previously used [49] with the number of iterations as 500 and the number of agents as 30. The algorithm is run 30 times and average and standard deviation on each test function are computed. Table 5 presents the test results of ANA, DA, WOA, and SSA on CEC-06 2019 test functions [49].

From Table 5, it can be concluded that each test function of the ANA algorithm on CEC functions is minimized towards one except on CEC01 and CEC06 that did not provide any convenient results; the runtime for these two functions was considerably long for taking results. ANA outperformed all the other algorithms on all the other test cases. This is another indicator of the ANA algorithm's outstanding performance and efficiency. It is worth mentioning that the WOA algorithm has the same result as ANA on the CEC03 function, but the value of the WOA algorithm in Table 5 has been estimated to only 4 decimal places. However, the standard deviation of WOA is 0.0 on the CEC03 function, which means WOA provides the same result each time it is run without any chances for further improvement.

**Table 5.** ANA, DA, WOA, and SSA test results on CEC-C06 2019 benchmark functions [49].

| Test Function | ANA | | DA [49] | | WOA [49] | | SSA [49] | |
|---|---|---|---|---|---|---|---|---|
| | Mean | Standard Deviation | Mean | Standard Deviation | Mean | Standard Deviation | Mean | Standard Deviation |
| CEC01 | - | - | $5.43 \times 10^{10}$ | $6.69 \times 10^{10}$ | $4.11 \times 10^{10}$ | $5.42 \times 10^{10}$ | **$6.05 \times 10^9$** | $4.75 \times 10^9$ |
| CEC02 | **4** | $2.87 \times 10^{-14}$ | 78.0368 | 87.7888 | 17.3495 | 0.0045 | 18.3434 | 0.0005 |
| CEC03 | **13.70240422** | $2.01 \times 10^{-11}$ | 13.7026 | 0.0007 | **13.7024** | 0 | 13.7025 | 0.0003 |
| CEC04 | **38.50887822** | 10.07245727 | 344.356 | 414.098 | 394.675 | 248.563 | 41.6936 | 22.2191 |
| CEC05 | **1.224598709** | 0.114632394 | 2.5572 | 0.3245 | 2.7342 | 0.2917 | 2.2084 | 0.1064 |
| CEC06 | - | | 9.8955 | 1.6404 | 10.7085 | 1.0325 | **6.0798** | 1.4873 |
| CEC07 | **116.5962143** | 8.825046006 | 578.953 | 329.398 | 490.684 | 194.832 | 410.396 | 290.556 |
| CEC08 | **5.472814997** | 0.429461877 | 6.8734 | 0.5015 | 6.909 | 0.4269 | 6.3723 | 0.5862 |
| CEC09 | **2.000963996** | 0.00341781 | 6.0467 | 2.871 | 5.9371 | 1.6566 | 3.6704 | 0.2362 |
| CEC10 | **2.718281828** | $4.44 \times 10^{-16}$ | 21.2604 | 0.1715 | 21.2761 | 0.1111 | 21.04 | 0.078 |

### 5.3. Comparative Study

There are several measures and techniques for comparing the performance of algorithms. Considering the importance of reaching optimality in optimization, this part presents a comparative study on the average of global best solutions of ANA, DA, PSO, and GA algorithms and ANA, PSO, GPSO, EPSO, LNPSO, VC-PSO, and SO-PSO algorithms on the standard benchmark functions used for testing the ANA algorithm. The algorithms' average global best solution on the standard benchmark functions are taken from Tables 3 and 4, and each is ranked from 1 to 4 for Table 3: 1 for the algorithm that provided the minimum value/result on the function and 4 for the algorithm that provided the maximum result, and 1 to 7 for Table 4: 1 for the algorithm that provided the minimum value/result on the function and 7 for the algorithm that provided the maximum result. Table 6 presents the ranking of ANA, DA, PSO, and GA algorithms on the 16 standard benchmark functions, and Table 7 presents the total number of first, second, third, and fourth rankings of the algorithms. Moreover, Table 8 demonstrates the ranking of ANA, PSO, GPSO, EPSO, LNPSO, VC-PSO, and SO-PSO on the six standard benchmark functions, and Table 9 presents the total number of first, second, third, fourth, fifth, sixth, and seventh rankings of the algorithms.

**Table 6.** ANA, DA, PSO, and GA ranking on the standard benchmark functions.

| Test Function | ANA | DA | PSO | GA |
|---|---|---|---|---|
| F1 | 3 | 1 | 2 | 4 |
| F2 | 2 | 1 | 3 | 4 |
| F3 | 3 | 1 | 2 | 4 |
| F4 | 1 | 2 | 3 | 4 |
| F5 | 2 | 1 | 3 | 4 |
| F7 | 4 | 2 | 1 | 3 |
| F9 | 3 | 2 | 1 | 4 |
| F10 | 1 | 2 | 3 | 4 |
| F11 | 3 | 2 | 1 | 4 |
| F12 | 3 | 2 | 1 | 4 |
| F13 | 1 | 2 | 4 | 3 |
| F14 | 1 | 4 | 3 | 2 |
| F15 | 1 | 4 | 2 | 3 |
| F16 | 1 | 4 | 2 | 3 |
| F17 | 3 | 4 | 2 | 1 |
| F18 | 1 | 3 | 4 | 2 |

**Table 7.** ANA, DA, PSO, and GA total number of ranking on the standard benchmark functions.

| Rank | ANA | DA | PSO | GA |
|---|---|---|---|---|
| First | 7 | 4 | 4 | 1 |
| Second | 2 | 7 | 5 | 2 |
| Third | 6 | 1 | 5 | 4 |
| Fourth | 1 | 4 | 2 | 9 |

**Table 8.** ANA, PSO, GPSO, EPSO, LNPSO, VC-PSO, and SO-PSO ranking on the standard benchmark functions.

| Test Function | ANA | PSO | GPSO | EPSO | LNPSO | VC-PSO | SO-PSO |
|---|---|---|---|---|---|---|---|
| F1 | 1 | 6 | 4 | 6 | 4 | 2 | 3 |
| F5 | 6 | 7 | 5 | 4 | 1 | 2 | 3 |
| F7 | 6 | 7 | 4 | 1 | 3 | 2 | 5 |
| F9 | 1 | 6 | 3 | 5 | 7 | 4 | 2 |
| F10 | 7 | 6 | 3 | 4 | 4 | 2 | 1 |
| F11 | 7 | 6 | 3 | 5 | 4 | 1 | 2 |

**Table 9.** ANA, PSO, GPSO, EPSO, LNPSO, VC-PSO, and SO-PSO total number of ranking on the standard benchmark functions.

| Rank | ANA | PSO | GPSO | EPSO | LNPSO | VC-PSO | SO-PSO |
|---|---|---|---|---|---|---|---|
| First | 2 | 0 | 0 | 1 | 1 | 1 | 1 |
| Second | 0 | 0 | 0 | 0 | 0 | 4 | 2 |
| Third | 0 | 0 | 3 | 0 | 1 | 0 | 2 |
| Fourth | 0 | 0 | 2 | 2 | 3 | 1 | 0 |
| Fifth | 0 | 0 | 1 | 2 | 0 | 0 | 1 |
| Sixth | 2 | 4 | 0 | 1 | 0 | 0 | 0 |
| Seventh | 2 | 2 | 0 | 0 | 1 | 0 | 0 |

Tables 6 and 7 demonstrate that the ANA algorithm has the highest first ranking number with a total of seven and the lowest fourth ranking with a total of only one in comparison to the famous DA, PSO, and GA algorithms. Furthermore, ANA demonstrates its efficiency once again by achieving the highest first rank on the six standard benchmark functions in comparison to PSO, GPSO, EPSO, LNPSO, VC-PSO, and SO-PSO algorithms as can be seen from Tables 8 and 9.

For providing a more detailed evaluation of the algorithm, ANA is ranked on the standard benchmark functions both by type of benchmark function and in total in comparison to DA, PSO, and GA algorithms. Table 10 presents ANA rankings on standard benchmark functions by type and in total. As demonstrated, the ranking by the type of the benchmark function for unimodal and multimodal test functions both are 2.50 and is 1.33 for composite test functions. Furthermore, if the global average performance of ANA is rounded to the nearest integer, then ANA ranks second amongst the four algorithms and the evaluated 16 benchmark functions. That demonstrates the algorithm's great per-formance compared to these famous algorithms. However, it is worth noting that no algorithm can perform the best for all optimization problems. Some algorithms will perform very well on some problems, while others will not perform as well as that algorithm [63].

**Table 10.** ANA ranking on standard benchmark functions by type and in total.

| Test function Type | Total Ranking | Total Ranking/No. of Function | Ranking (1–4) |
|---|---|---|---|
| Unimodal | 15 | 15/6 | 2.50 |
| Multimodal | 10 | 10/4 | 2.50 |
| Composite | 8 | 8/6 | 1.33 |
| Total | 33 | 33/16 | 2.06 |

### 5.4. ANA versus FDO

For providing a more detailed insight into the ANA algorithm and proving its significant performance, it has been compared to another novel algorithm that has an outstanding performance. The 16 standard benchmark functions used for testing the ANA algorithm have been selected, and the results have been compared to the FDO algorithm. FDO has been selected to conduct the comparison for three main reasons. First, FDO is a PSO-based algorithm same as ANA. Second, the FDO algorithm's both outstanding performance and outperforming all the standard GA, PSO, DA, WOA, and SSA algorithms have been proven in this work [49]. Third, the algorithm's implementation is publicly provided by the authors.

Regarding the algorithms' parameter settings, both the algorithms have been run with 30 search agents and 500 iterations 30 times. The *wf*, which is the FDO's single parameter, has been set to 0 in all the test cases. The global best agent for each turn has been recorded, and the results are demonstrated in box and whisker plots in Figure 6a–p. Figure 6 is another indicator for the ANA algorithm's significant performance since the results show the algorithm's outperforming FDO on seven test functions that are namely F2, F4, F5, F11, F12, F14, and F17 and very comparative results on the others. Table A5 in Appendix A contains the 30 times test results of ANA and FDO algorithms on the standard benchmark functions.

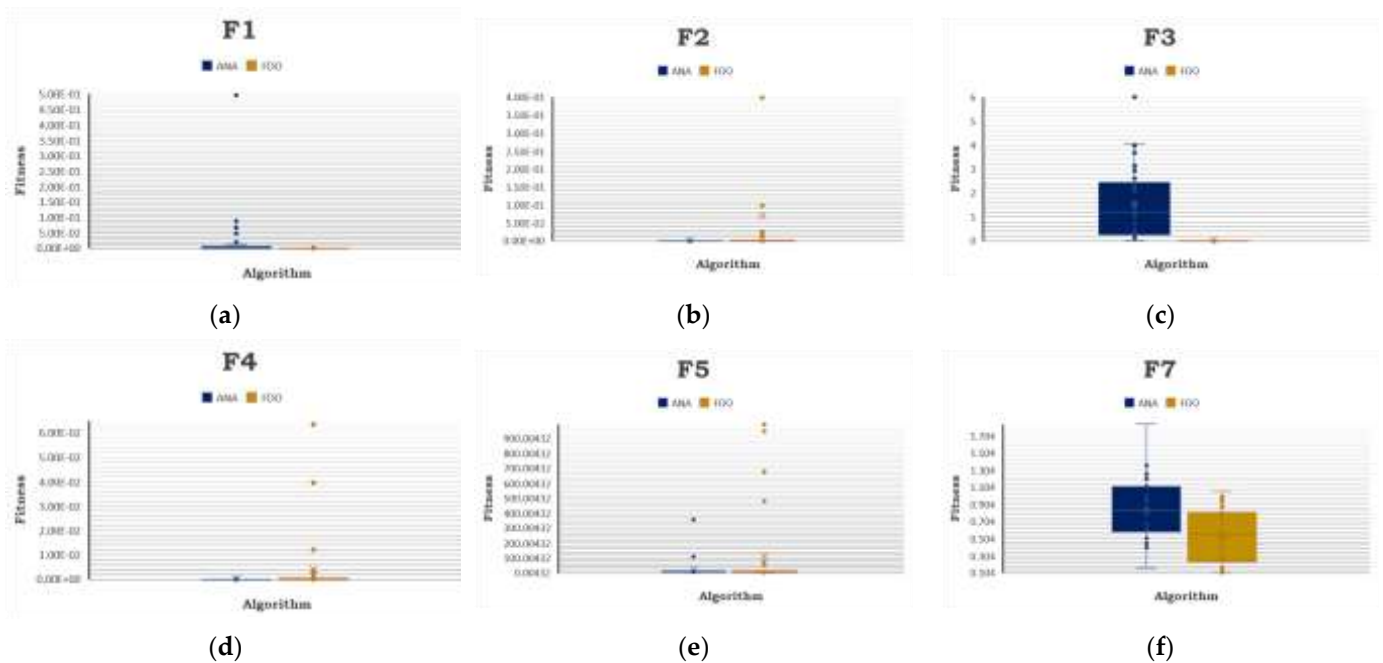

(a)   (b)   (c)

(d)   (e)   (f)

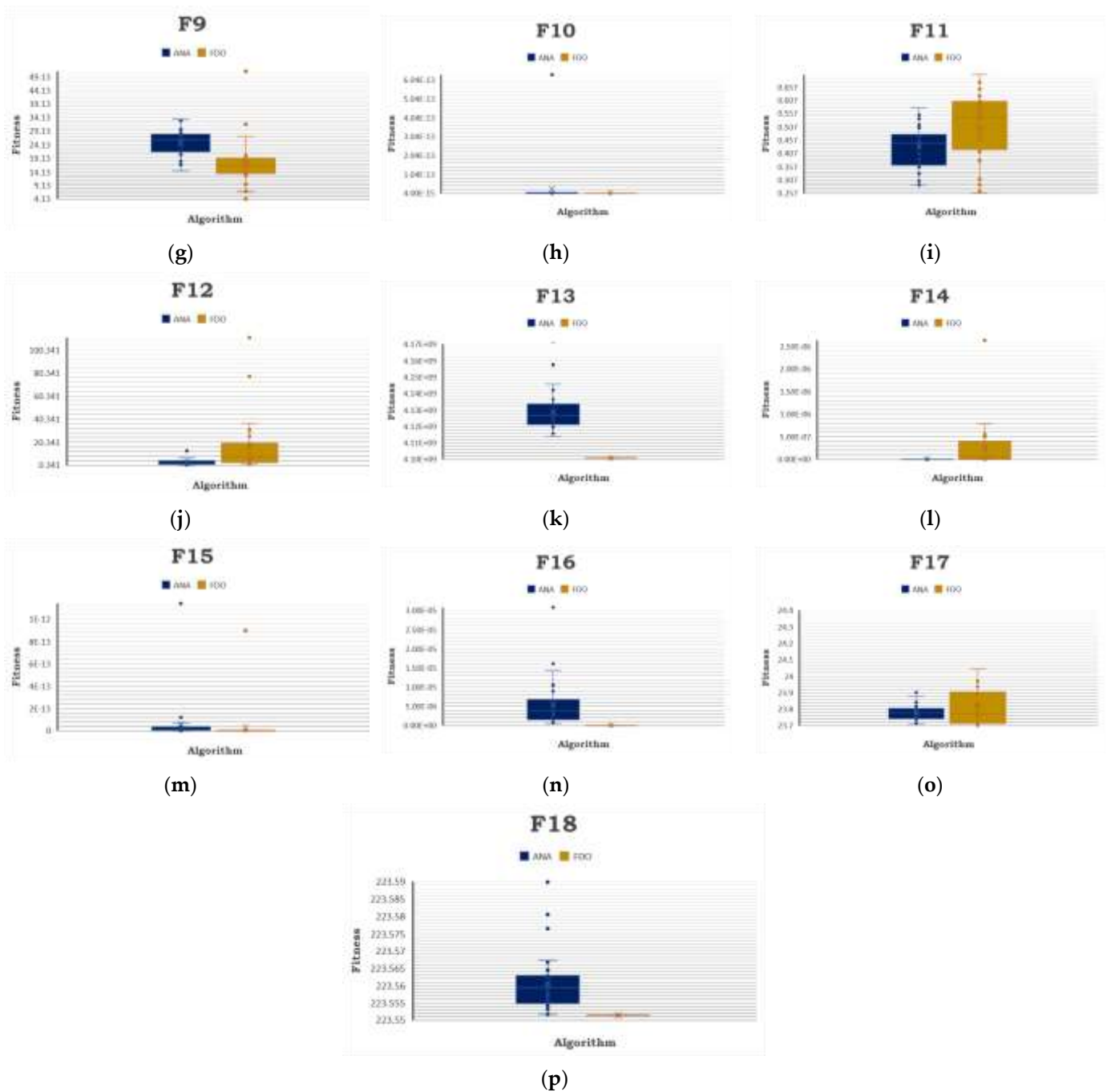

**Figure 6.** Box and whisker plot of ANA and FDO on the standard benchmark functions: (**a**) ANA versus FDO on F1; (**b**) ANA versus FDO on F2; (**c**) ANA versus FDO on F3; (**d**) ANA versus FDO on F4; (**e**) ANA versus FDO on F5; (**f**) ANA versus FDO on F7; (**g**) ANA versus FDO on F9; (**h**) ANA versus FDO on F10; (**i**) ANA versus FDO on F11; (**j**) ANA versus FDO on F12; (**k**) ANA versus FDO on F13; (**l**) ANA versus FDO on F14; (**m**) ANA versus FDO on F15; (**n**) ANA versus FDO on F16; (**o**) ANA versus FDO on F17; (**p**) ANA versus FDO on F18.

### 5.5. Statistical Test

To show that the results presented in the previous section are statistically significant, the *p* values of the Student's *t*-test, Welch's *t*-test, and Wilcoxon signed-rank test are found for all the standard benchmark functions, and the results are shown in Table 11. In Table 11, the comparison is conducted only between ANA and FDO algorithms because FDO was already tested against DA, PSO, and GA algorithms in this paper [49]. According to the mentioned work, it has been proven that the FDO results are statistically significant compared to DA, PSO, and GA algorithms.

As shown in Table 11, ANA results are considered significant in 10 statistical tests of Student's *t*-test that are namely F3, F4, F5, F7, F9, F11, F12, F13, F16, and F18, 7 statistical tests of Welch's *t*-test that are namely F1, F2, F4, F5, F10, F15, and F17, and 3 statistical tests of Wilcoxon signed-rank test that are namely F2, F5, and F17 that is because the results are less than 0.05. According to this work [64], the statistical test results of the Wilcoxon signed-rank test should be relied on for ensuring the statistical significance of the ANA algorithm in comparison to FDO as the data is not normally distributed nor homoscedastic. Tables A6 and A7 in Appendix A contains the normality test using the Shapiro–Wilk test and the homoscedastic test using Levene's test of ANA and FDO on the standard benchmark functions.

**Table 11.** The *p* values of the Student's t-test, Welch's t-test, and Wilcoxon signed-rank test of ANA and FDO on the standard benchmark functions.

| Test Function | Student's *t*-test | Welch's t-test | Wilcoxon signed-rank test |
|---|---|---|---|
| F1 | 0.066202 | **0.137803** | $1.86 \times 10^{-9}$ |
| F2 | 0.092068 | **0.189324** | **0.685047** |
| F3 | **0.00001** | $3.0678 \times 10^{-6}$ | $1.86 \times 10^{-9}$ |
| F4 | **0.046755** | 0.0988506 | $3.73 \times 10^{-9}$ |
| F5 | **0.04671** | 0.0976895 | **0.404495** |
| F7 | **0.000517** | 0.00104329 | 0.00761214 |
| F9 | **0.00021** | $5.827 \times 10^{-5}$ | $1.061 \times 10^{-5}$ |
| F10 | 0.5 | **0.295841** | 0.00559672 |
| F11 | **0.001898** | 0.00403645 | 0.00322299 |
| F12 | **0.001298** | 0.00372609 | $5.1446 \times 10^{-6}$ |
| F13 | **0.00001** | $1.95 \times 10^{-13}$ | $1.2508 \times 10^{-6}$ |
| F14 | - | 0.0120538 | $3.73 \times 10^{-9}$ |
| F15 | 0.5 | **0.536272** | 0.00021761 |
| F16 | **0.00001** | $4.0245 \times 10^{-5}$ | $1.86 \times 10^{-9}$ |
| F17 | 0.100256 | **0.203807** | **0.761065** |
| F18 | **0.00001** | $1.3746 \times 10^{-6}$ | $1.86 \times 10^{-9}$ |

* The bold *p* values of the tests indicate the significant results.

### 5.6. Real-World Applications of ANA

To prove the feasibility of the algorithm and evaluate its performance, ANA has been applied for solving two different real-world engineering problems.

#### 5.6.1. ANA on Aperiodic Antenna Array Design

The problem investigated for measuring the feasibility of the ANA algorithm is aperiodic antenna array design. In today's technological society, developing products that are more efficient and economical than their predecessors is quite crucial and highly demanded. The development of radar technology is a reason among several ones that led to the huge demand for innovation in the area of antenna array design [65]. Over the years, several various antennas have been designed to accommodate the industry's growing needs. One major requirement for antenna design has been the ability to position the elements of non-uniform arrays to obtain the peak sidelobe level (SLL).

Design techniques for array element placement include thinning, numerical optimization, and other methods. However, even with modern tools, the designing problem remains computationally challenging. For reducing the complexity of the design, and the optimization algorithm is used for optimizing element positions and minimizing peak sidelobe levels [66].

In non-uniform isotropic arrays, there are 10 elements, and only four of them need to be optimized. Thus, the application is a four-dimensional optimization problem. It is worth noting that designing the aperiodic antenna array is a convex problem since all the lines joining every two elements/points lie in the set, and minimization algorithms are considered to be quite efficient in optimizing this problem and reaching the optimum is much likely [67]. For more details on this problem, interested readers may refer to this paper [68]. Figure 7 demonstrates an array configuration for 10 elements.

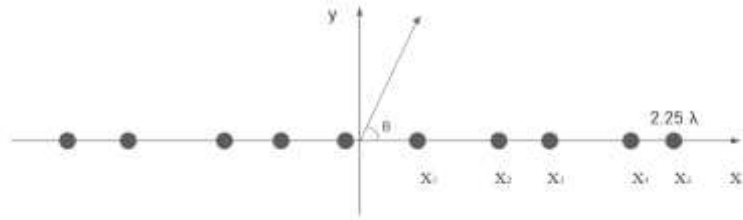

**Figure 7.** Configuration of an array with 10 elements.

The fitness function of the problem is described by the following expression:

$$f = max\{20\log|AF(\theta)|\} \tag{9}$$

where,

$$AF(\theta) = \sum_{i=1}^{4} \cos[2\pi x_i(\cos\theta - \cos\theta_s)] + \cos[2.25 \times 2\pi(\cos\theta - \cos\theta_s)] \tag{10}$$

Consider $\theta_s = 90°$ [68]

For achieving an improved peak SLL in non-uniform arrays, the element positions need to be optimized in a real number vector. In addition to that, to mitigate grating lobe level, a certain element spacing limit is required. Equation (11) shows the constraints of the problem.

$$x_i \in (0, 2.25)|x_i - x_j| > 0.25\lambda_0$$
$$min\{x_i\} > 0.125\lambda_0.\ i = 1,2,3,4.\ i \neq j \tag{11}$$

ANA algorithm is applied for optimizing this problem concerning the constraints mentioned in Equation (11). Here, 20 search agents and 200 iterations are used. The algorithm reached its optimum solution in iteration 57 with element positions {1.5959, 0.3081, 0.8747, 0.6072}.

### 5.6.2. ANA on Frequency-Modulated Synthesis

A frequency-modulated synthesis is a form of sound synthesis whereby the frequency of a waveform is changed by modulating its frequency. It is a complex real-world engineering optimization problem that has a fundamental role in several modern music systems. For the FM synthesis to create a harmonic sound, six parameters need to be optimized. Thus, the parameter optimization of an FM synthesizer is a six-dimensional optimization problem where the vector to be optimized is $\vec{x} = \{a_1, w_1, a_2, w_2, a_3, w_3\}$ of the sound wave given in Equation (12). The objective of this problem is to generate a sound in Equation (12) that is similar to the target sound in Equation (13).

$$y(t) = a_1.\sin(w_1.t.+a_2.\sin(w_2.t.\theta + a_3.\sin(w_3.t.\theta))) \tag{12}$$

$$y_o(t) = (1.0).\sin\left((5.0).t.+(1.5).\sin\left((4.8).t.\theta + (2.0).\sin\left((4.9).t.\theta\right)\right)\right) \tag{13}$$

respectively, where the parameters are defined in the range [−6.4, 6.35] and $\theta = 2\pi/100$. The fitness function can be calculated using Equation (14), which is the summation of square errors between the estimated wave, i.e., the result of Equation (12) and the target wave in Equation (13), while $t = 100$ turns. Interested readers can find more details on this problem in this work by [69].

$$f(\vec{x}) = \sum_{t=0}^{100} (y(t) - y_o(t))^2 \qquad (14)$$

ANA is applied to optimize this with 30 search agents and 200 iterations. The global best value converges to near-global optimal value from iteration 45 with parameters $\vec{x} = \{a_1 = -0.1253, w_1 = 3.0527, a_2 = -5.2198, w_2 = -4.4465, a_3 = 2.5656, w_3 = 2.8532\}$.

## 6. Conclusions

A novel swarm intelligent algorithm for optimizing single objective problems called the ant nesting algorithm was proposed. The proposed algorithm is inspired by Leptothorax ant behavior when dropping grains for constructing a new nest. The algorithm mimics what a swarm of worker ants do while searching for a position to drop grain around their queen ant in addition to a model of their random walk performance while searching. ANA employs slope-side lengths of Pythagorean theorem from the difference between previous and current deposition positions to the best local deposition position as one side and their fitness values as the other side to generate weights that drive the search agents towards optimality. In addition to that, ANA depends on the randomization mechanism in all the phases of searching including initialization, exploration, and exploitation.

Regarding ANA's performance after testing on several standards and modern test functions and real-world applications, it has been discovered that the common parameter settings like several search agents and iterations have a great impact on the algorithm's performance as is the case with the other metaheuristic algorithms. Increasing the number of search agents and iterations may provide much better and more accurate results, and the reverse is also true. However, on the other hand, increasing the number of search agents and/or iterations means utilizing more resources. Thus, a wise choice needs to be made while setting the parameters.

This paper is only providing a new method for reaching optimality in single-objective problems and several works can be suggested for conducting further studies. The development of binary and multi-objective versions of the ANA algorithm is one recommendation for solving a greater range of diverse optimization problems. Another recommendation is integrating more evolutionary operators into the algorithm for improving performance and/or more effective utilization of resources. In addition to that, hybridizing the algorithm with the other algorithms and using their features is another exceptional suggestion. Finally, the most important is applying the algorithm in solving real-world applications. It is strongly recommended that the algorithm is used for solving problems practically.

**Author Contributions:**

Conceptualization, Deeam Najmadeen Hama Rashid; methodology, Deeam Najmadeen Hama Rashid; software, Deeam Najmadeen Hama Rashid; validation, Deeam Najmadeen Hama Rashid; formal analysis, Deeam Najmadeen Hama Rashid, Tarik A. Rashid and Seyedali Mirjalili; investigation, Tarik A. Rashid and Seyedali Mirjalili; resources, Deeam Najmadeen Hama Rashid and Tarik A. Rashid; data curation, Deeam Najmadeen Hama Rashid; writing—original draft preparation, Deeam Najmadeen Hama Rashid; writing—review and editing, Deeam Najmadeen Hama Rashid, Tarik A. Rashid and Seyedali Mirjalili; visualization, Deeam Najmadeen Hama Rashid; supervision, Tarik A. Rashid; project administration, Deeam Najmadeen Hama Rashid and Tarik A. Rashid; funding acquisition, Tarik A. Rashid. All authors have read and agreed to the published version of the manuscript.

**Funding:**

This research received no external funding.

**Conflicts of Interest:**

The authors declare no conflict of interest.

## Appendix A

<div align="center">

**Table A1.** Unimodal test functions [38].

</div>

| Function | Dimension | Range | Shift position | $f_{min}$ |
|---|---|---|---|---|
| $F1(x) = \sum\limits_{i=1}^{n} x_i^2$ | 10 | $[-100, 100]$ | $[-30, -30, \dots -30]$ | 0 |
| $F2(x) = \sum\limits_{i=1}^{n} \lvert x_i \rvert + \prod\limits_{i=1}^{n} \lvert x_i \rvert$ | 10 | $[-10, 10]$ | $[-3, -3, \dots -3]$ | 0 |
| $F3(x) = \sum\limits_{i=1}^{n} \left( \sum\limits_{j-1}^{i} x_j \right)^2$ | 10 | $[-100, 100]$ | $[-30, -30, \dots -30]$ | 0 |
| $F4(x) = \max\limits_{i}\{\lvert x \rvert, 1 \leq i \leq n\}$ | 10 | $[-100, 100]$ | $[-30, -30, \dots -30]$ | 0 |
| $F5(x) = \sum\limits_{i=1}^{n-1}[100(x_{i+1} - x_i^2)^2 + (x_i - 1)^2]$ | 10 | $[-30, 30]$ | $[-15, -15, \dots -15]$ | 0 |
| $F6(x) = \sum\limits_{i=1}^{n}([x_i + 0.5])^2$ | 10 | $[-100, 100]$ | $[-750, \dots -750]$ | 0 |
| $F7(x) = \sum\limits_{i=1}^{n} i x_i^4 + \text{random}[0, 1]$ | 10 | $[-1.28, 1.28]$ | $[-0.25, \dots -0.25]$ | 0 |

<div align="center">

**Table A2.** Multimodal test functions [38].

</div>

| Function | Dimension | Range | Shift position | $f_{min}$ |
|---|---|---|---|---|
| $F8(x) = \sum\limits_{i=1}^{n} -x_i^2 \sin\left(\sqrt{\lvert x_i \rvert}\right)$ | 10 | $[-500, 500]$ | $[-300, \dots -300]$ | $-418.9829$ |
| $F9(x) = \sum\limits_{i=1}^{n}[x_i^2 - 10\cos(2\pi x_i) + 10]$ | 10 | $[-5.12, 5.12]$ | $[-2, -2, \dots -2]$ | 0 |
| $F10(x) = -20\,exp\left(-0.2\sqrt{\sum\limits_{i=1}^{n} x_i^2}\right)$ $-\,exp\left(\frac{1}{n}\sum\limits_{i=1}^{n}\cos(2\pi x_i)\right) + 20 + e$ | 10 | $[-32, 32]$ | $[0, 0, \dots 0]$ | 0 |
| $F11(x) = \frac{1}{4000}\sum\limits_{i=1}^{n} x_i^2 - \prod\limits_{i=1}^{n}\cos\left(\frac{x_i}{\sqrt{i}}\right) + 1$ | 10 | $[-600, 600]$ | $[-400, \dots -400]$ | 0 |
| $F12(x) = \frac{\pi}{n}\{10\sin(\pi y_1) + \sum_{i=1}^{n-1}(y_i - 1)^2[1 + 10\sin^2(\pi y_{i+1})] + (y_n - 1)^2\} + \sum_{i=1}^{n} u(x_i, 10, 100, 4).$ $y_i = 1 + \frac{x+1}{4}.$ $u(x_i, a, k, m) = \begin{cases} k(x_i - a)^m & x_i > a \\ 0 & -a < x_i < a \\ k(-x_i - a)^m & x_i < -a \end{cases}$ | 10 | $[-50, 50]$ | $[-30, 30, \dots 30]$ | 0 |

**Table A3.** Composite test functions [38].

| Function | Dimension | Range | $f_{min}$ |
|---|---|---|---|
| **F13 (CF1)**<br>$f1, f2, f3 \dots f10 =$ Sphere function<br>$\delta1, \delta2, \delta3 \dots \delta10 = [1,1,1, \dots .1]$<br>$\lambda1, \lambda2, \lambda3 \dots \lambda10 = \left[\dfrac{5}{100}, \dfrac{5}{100}, \dfrac{5}{100}, \dots \dfrac{5}{100}\right]$ | 10 | $[-5, 5]$ | 0 |
| **F14 (CF2)**<br>$f1, f2, f3 \dots f10 =$ Griewank's function<br>$\delta1, \delta2, \delta3 \dots \delta10 = [1,1,1, \dots .1]$<br>$\lambda1, \lambda2, \lambda3 \dots \lambda10 = \left[\dfrac{5}{100}, \dfrac{5}{100}, \dfrac{5}{100}, \dots \dfrac{5}{100}\right]$ | 10 | $[-5, 5]$ | 0 |
| **F15 (CF3)**<br>$f1, f2, f3 \dots f10 =$ Griewank's function<br>$\delta1, \delta2, \delta3 \dots \delta10 = [1,1,1, \dots .1]$<br>$\lambda1, \lambda2, \lambda3 \dots \lambda10 = [1,1,1, \dots .1]$ | 10 | $[-5, 5]$ | 0 |
| **F16 (CF4)**<br>$f1, f2 =$ Ackley's function<br>$f3, f4 =$ Rastrigin's function<br>$f5, f6 =$ Weierstrass function<br>$f7, f8 =$ Griewank's function<br>$f9, f10 =$ Sphere function<br>$\delta1, \delta2, \delta3 \dots \delta10 = [1,1,1, \dots .1]$<br>$\lambda1, \lambda2, \lambda3 \dots \lambda10 = \left[\dfrac{5}{32}, \dfrac{5}{32}, 1, 1, \dfrac{5}{0.5}, \dfrac{5}{0.5}, \dfrac{5}{100}, \dfrac{5}{100}, \dfrac{5}{100}, \dfrac{5}{100}\right]$ | 10 | $[-5, 5]$ | 0 |
| **F17 (CF5)**<br>$f1, f2 =$ Rastrigin's function<br>$f3, f4 =$ Weierstrass function<br>$f5, f6 =$ Griewank's function<br>$f7, f8 =$ Ackley's function<br>$f9, f10 =$ Sphere function<br>$\delta1, \delta2, \delta3 \dots \delta10 = [1,1,1, \dots .1]$<br>$\lambda1, \lambda2, \lambda3 \dots \lambda10 = \left[\dfrac{1}{5}, \dfrac{1}{5}, \dfrac{5}{0.5}, \dfrac{5}{0.5}, \dfrac{5}{100}, \dfrac{5}{100}, \dfrac{5}{32}, \dfrac{5}{32}, \dfrac{5}{100}, \dfrac{5}{100}\right]$ | 10 | $[-5, 5]$ | 0 |
| **F18 (CF6)**<br>$f1, f2 =$ Rastrigin's function<br>$f3, f4 =$ Weierstrass function<br>$f5, f6 =$ Griewank's function<br>$f7, f8 =$ Ackley's function<br>$f9, f10 =$ Sphere function<br>$\delta1, \delta2, \delta3 \dots \delta10 = [0.1,0.2,0.3, 0.4,0.5,0.6,0.7,0.8,0.9,1]$<br>$\lambda1, \lambda2, \lambda3 \dots \lambda10 = \left[0.1*\dfrac{1}{5}, 0.2*\dfrac{1}{5}, 0.3*\dfrac{5}{0.5}, 0.4*\dfrac{5}{0.5}, 0.5*\dfrac{5}{100}, 0.6 \right.$<br>$\left. *\dfrac{5}{100}, 0.7*\dfrac{5}{32}, 0.8*\dfrac{5}{32}, 0.9*\dfrac{5}{100}, 1*5/100\right]$ | 10 | $[-5, 5]$ | 0 |

**Table A4.** CEC-C06 2019 test functions [62].

| Function No. | Function Name | Dimension | Range | $f_{min}$ |
|---|---|---|---|---|
| CEC01 | STORN'S CHEBYSHEV POLYNOMIAL FITTING PROBLEM | 9 | [−8192, 8192] | 1 |
| CEC02 | INVERSE HILBERT MATRIX PROBLEM | 16 | [−16384, 16384] | 1 |
| CEC03 | LENNARD-JONES MINIMUM ENERGY CLUSTER | 18 | [−4, 4] | 1 |
| CEC04 | RASTRIGIN'S FUNCTION | 10 | [−100, 100] | 1 |
| CEC05 | GRIEWANGK'S FUNCTION | 10 | [−100, 100] | 1 |
| CEC06 | WEIERSTRASS FUNCTION | 10 | [−100, 100] | 1 |
| CEC07 | MODIFIED SCHWEFEL'S FUNCTION | 10 | [−100, 100] | 1 |
| CEC08 | EXPANDED SCHAFFER'S F6 FUNCTION | 10 | [−100, 100] | 1 |
| CEC09 | HAPPY CAT FUNCTION | 10 | [−100, 100] | 1 |
| CEC10 | ACKLEY FUNCTION | 10 | [−100, 100] | 1 |

**Table A5.** ANA and FDO test results of 30 turns on standard benchmark functions.

| Turn | F1 ANA | F1 FDO | F2 ANA | F2 FDO | F3 ANA | F3 FDO | F4 ANA | F4 FDO | F5 ANA | F5 FDO | F7 ANA | F7 FDO |
|---|---|---|---|---|---|---|---|---|---|---|---|---|
| 1 | $1.11 \times 10^{-4}$ | $3.79 \times 10^{-29}$ | $8.86 \times 10^{-5}$ | $1.95 \times 10^{-8}$ | 2.391747144 | $1.69 \times 10^{-11}$ | $7.11 \times 10^{-15}$ | $1.64 \times 10^{-8}$ | 7.432984162 | 5.389371864 | 0.866804261 | 0.251164251 |
| 2 | $7.23 \times 10^{-8}$ | $8.58 \times 10^{-28}$ | $3.54 \times 10^{-5}$ | 0.025049009 | 3.676873939 | $2.43 \times 10^{-13}$ | 0 | $1.27 \times 10^{-12}$ | 8.133632758 | 5.212061321 | 0.165221295 | 0.154350971 |
| 3 | 0.048192575 | $1.14 \times 10^{-28}$ | $1.50 \times 10^{-4}$ | $4.08 \times 10^{-6}$ | 2.600793805 | $2.85 \times 10^{-9}$ | 0 | 0.063337923 | 7.689508303 | 480.5348286 | 1.202136818 | 0.186249902 |
| 4 | $2.33 \times 10^{-5}$ | $8.84 \times 10^{-28}$ | $2.28 \times 10^{-5}$ | 0.031798849 | 3.977493658 | $5.92 \times 10^{-13}$ | 0 | 0.004086299 | 8.652707669 | 6.963351041 | 1.026315594 | 0.986077172 |
| 5 | $3.52 \times 10^{-6}$ | $3.27 \times 10^{-24}$ | $4.25 \times 10^{-5}$ | $2.22 \times 10^{-15}$ | 0.153875296 | $4.75 \times 10^{-9}$ | 0 | $6.39 \times 10^{-7}$ | 6.566481526 | 7.547774394 | 0.617639517 | 0.69199572 |
| 6 | 0.066742805 | $4.92 \times 10^{-28}$ | $2.41 \times 10^{-5}$ | $2.34 \times 10^{-12}$ | 0.023104804 | $5.44 \times 10^{-11}$ | 0 | 0.01211212 | 7.703774666 | 946.2636218 | 0.790594357 | 0.108666552 |
| 7 | 0.003550881 | $4.92 \times 10^{-28}$ | $1.64 \times 10^{-5}$ | $8.33 \times 10^{-6}$ | 1.065088711 | $1.55 \times 10^{-11}$ | 0 | $5.52 \times 10^{-9}$ | 18.36487837 | 4.151222988 | 1.357209894 | 0.174620628 |
| 8 | $1.01 \times 10^{-5}$ | $2.52 \times 10^{-28}$ | $1.66 \times 10^{-5}$ | $2.17 \times 10^{-5}$ | 1.568160542 | $6.87 \times 10^{-14}$ | 0 | $7.77 \times 10^{-8}$ | 8.018400623 | 5.380822727 | 0.517424969 | 0.797722139 |
| 9 | $3.12 \times 10^{-7}$ | $1.72 \times 10^{-27}$ | $1.45 \times 10^{-4}$ | $1.48 \times 10^{-11}$ | 0.102410755 | $6.46 \times 10^{-13}$ | 0 | $1.06 \times 10^{-9}$ | 5.536287826 | 4.778696619 | 0.840359451 | 0.284824671 |
| 10 | $5.26 \times 10^{-6}$ | $6.31 \times 10^{-28}$ | $3.86 \times 10^{-5}$ | $2.01 \times 10^{-8}$ | 0.483393911 | $9.41 \times 10^{-12}$ | 0 | 0.001673928 | 8.546610565 | 3.215025725 | 0.810047666 | 0.405013575 |
| 11 | $2.78 \times 10^{-6}$ | $1.14 \times 10^{-28}$ | $3.67 \times 10^{-5}$ | $3.64 \times 10^{-14}$ | 0.864879054 | $1.42 \times 10^{-11}$ | 0 | $2.51 \times 10^{-10}$ | 7.874299055 | 0.004319718 | 1.117170129 | 0.935591962 |
| 12 | 0.019792811 | $3.45 \times 10^{-25}$ | $8.81 \times 10^{-6}$ | $2.22 \times 10^{-12}$ | 0.431912446 | $4.05 \times 10^{-12}$ | 0 | 0 | 7.893741339 | 7.583576095 | 0.953578507 | 0.104218667 |
| 13 | 0.013921899 | $1.77 \times 10^{-28}$ | $2.11 \times 10^{-6}$ | $3.29 \times 10^{-10}$ | 1.196097344 | $6.57 \times 10^{-11}$ | $9.24 \times 10^{-14}$ | $9.50 \times 10^{-9}$ | 8.207516834 | 6.766245897 | 0.832446794 | 0.873904075 |
| 14 | $2.47 \times 10^{-4}$ | $1.01 \times 10^{-25}$ | $1.04 \times 10^{-5}$ | $3.31 \times 10^{-10}$ | 1.215643383 | $6.47 \times 10^{-11}$ | 0 | $3.71 \times 10^{-6}$ | 8.370475137 | 5.806708272 | 0.417010569 | 0.445590624 |
| 15 | $1.14 \times 10^{-5}$ | $2.26 \times 10^{-25}$ | $4.63 \times 10^{-6}$ | 1.511854865 | 0.3075614 | $2.83 \times 10^{-12}$ | 0 | $8.99 \times 10^{-6}$ | 7.848728655 | 76.72611459 | 1.11353184 | 0.779961586 |
| 16 | 0.496882545 | $2.08 \times 10^{-25}$ | $5.05 \times 10^{-6}$ | $9.10 \times 10^{-14}$ | 4.031404303 | $3.80 \times 10^{-11}$ | $3.55 \times 10^{-15}$ | $1.41 \times 10^{-4}$ | 24.03502558 | 4.969049398 | 1.258567091 | 0.729519408 |
| 17 | $2.26 \times 10^{-5}$ | $8.84 \times 10^{-29}$ | $4.43 \times 10^{-5}$ | $2.68 \times 10^{-12}$ | 0.222310655 | $1.18 \times 10^{-10}$ | 0 | $9.79 \times 10^{-4}$ | 105.6739992 | 4.118419197 | 0.670713808 | 0.722250032 |
| 18 | $9.52 \times 10^{-5}$ | $7.83 \times 10^{-28}$ | $1.05 \times 10^{-4}$ | $1.16 \times 10^{-4}$ | 0.235561892 | $1.67 \times 10^{-9}$ | 0 | $1.32 \times 10^{-4}$ | 7.950295796 | 676.1226102 | 1.025029377 | 0.297330263 |
| 19 | $6.08 \times 10^{-6}$ | $8.64 \times 10^{-25}$ | $4.88 \times 10^{-5}$ | 0.398966573 | 2.271872396 | $2.33 \times 10^{-13}$ | 0 | $2.28 \times 10^{-8}$ | 355.4219987 | 6.054911494 | 1.214174323 | 0.171875021 |
| 20 | $3.59 \times 10^{-4}$ | $3.79 \times 10^{-28}$ | $6.07 \times 10^{-5}$ | $4.98 \times 10^{-9}$ | 2.107150179 | $5.82 \times 10^{-14}$ | $1.42 \times 10^{-14}$ | $6.19 \times 10^{-12}$ | 8.141655326 | 9.598117721 | 0.60764254 | 0.518382895 |
| 21 | $1.39 \times 10^{-6}$ | $1.64 \times 10^{-28}$ | $1.49 \times 10^{-5}$ | $1.74 \times 10^{-5}$ | 1.245802434 | $4.86 \times 10^{-14}$ | 0 | $6.26 \times 10^{-11}$ | 8.085746671 | 8.081052732 | 0.944769536 | 0.838040939 |
| 22 | $9.09 \times 10^{-6}$ | $1.07 \times 10^{-25}$ | $7.92 \times 10^{-6}$ | $1.54 \times 10^{-5}$ | 0.187935469 | $5.75 \times 10^{-14}$ | 0 | $7.33 \times 10^{-10}$ | 5.964614538 | 994.04003 | 0.536293549 | 0.539834343 |
| 23 | 0.027121243 | $9.72 \times 10^{-28}$ | $3.43 \times 10^{-5}$ | $5.24 \times 10^{-10}$ | 3.129001513 | $1.26 \times 10^{-10}$ | 0 | $2.13 \times 10^{-14}$ | 9.549734175 | 5.179255928 | 0.680443422 | 0.784514643 |
| 24 | $2.24 \times 10^{-5}$ | $1.06 \times 10^{-24}$ | $1.06 \times 10^{-5}$ | $5.98 \times 10^{-4}$ | 2.318721348 | $5.56 \times 10^{-9}$ | $1.07 \times 10^{-14}$ | $7.72 \times 10^{-5}$ | 8.193949903 | 3.072962007 | 0.447296236 | 0.887949921 |
| 25 | $4.07 \times 10^{-5}$ | $6.73 \times 10^{-24}$ | $1.07 \times 10^{-5}$ | $3.33 \times 10^{-14}$ | 0.121714672 | $3.85 \times 10^{-8}$ | $8.88 \times 10^{-12}$ | 0.039483383 | 14.59412606 | 5.38125078 | 1.142057744 | 0.494719509 |
| 26 | $1.02 \times 10^{-5}$ | $7.32 \times 10^{-28}$ | $7.53 \times 10^{-6}$ | 0.098319044 | 2.918241761 | $1.27 \times 10^{-11}$ | $1.07 \times 10^{-14}$ | $5.05 \times 10^{-8}$ | 8.276483401 | 9.063486988 | 0.399833358 | 0.986881941 |
| 27 | 0.088125129 | $5.05 \times 10^{-29}$ | $1.75 \times 10^{-5}$ | $1.16 \times 10^{-9}$ | 6.004686234 | $1.13 \times 10^{-11}$ | 0 | $3.55 \times 10^{-15}$ | 9.092436574 | 6.344508738 | 0.594489623 | 0.106760494 |
| 28 | $2.28 \times 10^{-4}$ | $3.79 \times 10^{-29}$ | $1.32 \times 10^{-5}$ | 0.013051787 | 1.397954863 | $2.31 \times 10^{-10}$ | 0 | 0.003278919 | 7.820602607 | 51.200028 | 0.509288229 | 0.745339406 |
| 29 | $7.52 \times 10^{-8}$ | $2.57 \times 10^{-23}$ | $1.78 \times 10^{-5}$ | $2.64 \times 10^{-13}$ | 0.739765656 | $6.68 \times 10^{-11}$ | 0 | $4.82 \times 10^{-7}$ | 4.912117549 | 7.903719687 | 1.844587 | 0.56002937 |
| 30 | $1.19 \times 10^{-5}$ | $3.22 \times 10^{-21}$ | $1.72 \times 10^{-5}$ | $7.28 \times 10^{-10}$ | 0.271081438 | $7.99 \times 10^{-8}$ | 0 | $4.64 \times 10^{-5}$ | 8.730117271 | 2.699807344 | 1.003968997 | 1.057408352 |

| | F9 | | F10 | | F11 | | F12 | | F13 | | F14 | |
|---|---|---|---|---|---|---|---|---|---|---|---|---|
| | ANA | FDO | ANA | FDO | ANA | ANA | ANA | FDO | ANA | FDO | ANA | FDO |
| 1 | 29.64993919 | 12.79952836 | $4.00 \times 10^{-15}$ | $4.00 \times 10^{-15}$ | 0.536669496 | $3.08 \times 10^{-35}$ | $3.08 \times 10^{-35}$ | 2.256208702 | $4.13 \times 10^{9}$ | $4.10 \times 10^{9}$ | $3.08 \times 10^{-35}$ | $1.05 \times 10^{-9}$ |
| 2 | 23.80324887 | 27.2700813 | $7.55 \times 10^{-15}$ | $4.00 \times 10^{-15}$ | 0.47016086 | $2.47 \times 10^{-33}$ | $2.47 \times 10^{-33}$ | 15.82823498 | $4.12 \times 10^{9}$ | $4.10 \times 10^{9}$ | $2.47 \times 10^{-33}$ | $9.27 \times 10^{-8}$ |
| 3 | 25.58357807 | 21.10470211 | $7.55 \times 10^{-15}$ | $4.00 \times 10^{-15}$ | 0.288038193 | $3.59 \times 10^{-26}$ | $3.59 \times 10^{-26}$ | 30.62229993 | $4.12 \times 10^{9}$ | $4.10 \times 10^{9}$ | $3.59 \times 10^{-26}$ | $1.76 \times 10^{-9}$ |
| 4 | 24.45921514 | 6.868915548 | $7.55 \times 10^{-15}$ | $4.00 \times 10^{-15}$ | 0.453501876 | $8.94 \times 10^{-34}$ | $8.94 \times 10^{-34}$ | 2.937099088 | $4.13 \times 10^{9}$ | $4.10 \times 10^{9}$ | $8.94 \times 10^{-34}$ | $4.98 \times 10^{-8}$ |
| 5 | 27.33640913 | 16.91917734 | $4.00 \times 10^{-15}$ | $7.55 \times 10^{-15}$ | 0.477649285 | 0 | | 18.26717959 | $4.14 \times 10^{9}$ | $4.10 \times 10^{9}$ | 0 | $3.63 \times 10^{-7}$ |
| 6 | 26.14245075 | 13.7293011 | $7.55 \times 10^{-15}$ | $4.00 \times 10^{-15}$ | 0.429930625 | 0 | | 20.72761646 | $4.13 \times 10^{9}$ | $4.10 \times 10^{9}$ | 0 | $4.88 \times 10^{-7}$ |
| 7 | 27.16351739 | 13.64136498 | $6.33 \times 10^{-13}$ | $4.00 \times 10^{-15}$ | 0.45735448 | 0 | | 3.707549452 | $4.13 \times 10^{9}$ | $4.10 \times 10^{9}$ | 0 | $6.45 \times 10^{-10}$ |
| 8 | 21.24192442 | 18.45987669 | $7.55 \times 10^{-15}$ | $4.00 \times 10^{-15}$ | 0.578880732 | $2.47 \times 10^{-34}$ | $2.47 \times 10^{-34}$ | 7.537494555 | $4.12 \times 10^{9}$ | $4.10 \times 10^{9}$ | $2.47 \times 10^{-34}$ | $7.47 \times 10^{-23}$ |
| 9 | 30.65835276 | 13.92959125 | $4.00\times10^{-15}$ | $4.00\times10^{-15}$ | 0.514372619 | 0 | 0 | 36.38507552 | $4.13\times10^9$ | $4.10\times10^9$ | 0 | $7.75\times10^{-7}$ |
| 10 | 24.77578868 | 18.32683402 | $4.00\times10^{-15}$ | $4.00\times10^{-15}$ | 0.330318389 | $8.49\times10^{-32}$ | $8.49\times10^{-32}$ | 25.16311633 | $4.12\times10^9$ | $4.10\times10^9$ | $8.49\times10^{-32}$ | $1.34\times10^{-7}$ |
| 11 | 18.03815799 | 13.92941673 | $4.00\times10^{-15}$ | $4.00\times10^{-15}$ | 0.519984074 | $3.08\times10^{-35}$ | $3.08\times10^{-35}$ | 1.053823092 | $4.12\times10^9$ | $4.10\times10^9$ | $3.08\times10^{-35}$ | $1.89\times10^{-7}$ |
| 12 | 17.42487553 | 16.33601874 | $7.55\times10^{-15}$ | $4.00\times10^{-15}$ | 0.340353647 | 0 | 0 | 15.64424504 | $4.13\times10^9$ | $4.10\times10^9$ | 0 | $5.56\times10^{-7}$ |
| 13 | 27.78577148 | 20.43700871 | $4.00\times10^{-15}$ | $4.00\times10^{-15}$ | 0.503012545 | 0 | 0 | 1.005512098 | $4.12\times10^9$ | $4.10\times10^9$ | 0 | $6.07\times10^{-7}$ |
| 14 | 30.55497212 | 19.95926201 | $4.00\times10^{-15}$ | $4.00\times10^{-15}$ | 0.362357712 | $4.13\times10^{-32}$ | $4.13\times10^{-32}$ | 110.9137614 | $4.16\times10^9$ | $4.10\times10^9$ | $4.13\times10^{-32}$ | $5.12\times10^{-9}$ |
| 15 | 33.47370848 | 18.94197091 | $4.00\times10^{-15}$ | $4.00\times10^{-15}$ | 0.405573309 | $1.23\times10^{-34}$ | $1.23\times10^{-34}$ | 3.343699564 | $4.13\times10^9$ | $4.10\times10^9$ | $1.23\times10^{-34}$ | $2.66\times10^{-10}$ |
| 16 | 27.57864812 | 7.016369273 | $7.55\times10^{-15}$ | $7.55\times10^{-15}$ | 0.550530947 | $6.97\times10^{-21}$ | $6.97\times10^{-21}$ | 18.58481895 | $4.12\times10^9$ | $4.10\times10^9$ | $6.97\times10^{-21}$ | $1.69\times10^{-11}$ |
| 17 | 22.07756327 | 16.91428894 | $4.00\times10^{-15}$ | $4.00\times10^{-15}$ | 0.452340051 | $7.70\times10^{-34}$ | $7.70\times10^{-34}$ | 77.1872071 | $4.12\times10^9$ | $4.10\times10^9$ | $7.70\times10^{-34}$ | $5.02\times10^{-7}$ |
| 18 | 28.19367857 | 16.14777042 | $4.00\times10^{-15}$ | $4.00\times10^{-15}$ | 0.355658197 | $4.01\times10^{-34}$ | $4.01\times10^{-34}$ | 3.79085804 | $4.13\times10^9$ | $4.10\times10^9$ | $4.01\times10^{-34}$ | $2.04\times10^{-8}$ |
| 19 | 27.12853068 | 51.1308107 | $4.00\times10^{-15}$ | $4.00\times10^{-15}$ | 0.303665389 | $1.23\times10^{-34}$ | $1.23\times10^{-34}$ | 11.27948275 | $4.13\times10^9$ | $4.10\times10^9$ | $1.23\times10^{-34}$ | $2.58\times10^{-7}$ |
| 20 | 27.03229743 | 13.73732476 | $1.47\times10^{-14}$ | $4.00\times10^{-15}$ | 0.434220836 | $2.45\times10^{-28}$ | $2.45\times10^{-28}$ | 3.201621381 | $4.12\times10^9$ | $4.10\times10^9$ | $2.45\times10^{-28}$ | $6.04\times10^{-33}$ |
| 21 | 23.98052652 | 12.93481639 | $7.55\times10^{-15}$ | $4.00\times10^{-15}$ | 0.453990137 | $2.37\times10^{-29}$ | $2.37\times10^{-29}$ | 2.24235298 | $4.17\times10^9$ | $4.10\times10^9$ | $2.37\times10^{-29}$ | $5.78\times10^{-8}$ |
| 22 | 30.48681497 | 16.18708655 | $7.55\times10^{-15}$ | $4.00\times10^{-15}$ | 0.448005501 | $1.43\times10^{-30}$ | $1.43\times10^{-30}$ | 18.02894669 | $4.12\times10^9$ | $4.10\times10^9$ | $1.43\times10^{-30}$ | $5.55\times10^{-7}$ |
| 23 | 20.61803459 | 18.90882319 | $7.55\times10^{-15}$ | $4.00\times10^{-15}$ | 0.366715922 | 0 | 0 | 7.29987667 | $4.13\times10^9$ | $4.10\times10^9$ | 0 | $4.66\times10^{-27}$ |
| 24 | 21.55669391 | 13.93157013 | $4.00\times10^{-15}$ | $4.00\times10^{-15}$ | 0.443166004 | $3.08\times10^{-35}$ | $3.08\times10^{-35}$ | 2.820756579 | $4.13\times10^9$ | $4.10\times10^9$ | $3.08\times10^{-35}$ | $7.95\times10^{-8}$ |
| 25 | 20.45554822 | 4.13272208 | $4.00\times10^{-15}$ | $4.00\times10^{-15}$ | 0.36216298 | 0 | 0 | 4.134924531 | $4.15\times10^9$ | $4.10\times10^9$ | 0 | $6.67\times10^{-9}$ |
| 26 | 16.73090762 | 9.52312913 | $4.00\times10^{-15}$ | $4.00\times10^{-15}$ | 0.477300194 | $9.00\times10^{-33}$ | $9.00\times10^{-33}$ | 1.897965144 | $4.13\times10^9$ | $4.10\times10^9$ | $9.00\times10^{-33}$ | $8.84\times10^{-10}$ |
| 27 | 21.69321924 | 20.66968085 | $4.00\times10^{-15}$ | $4.00\times10^{-15}$ | 0.385521001 | $8.35\times10^{-30}$ | $8.35\times10^{-30}$ | 20.48011935 | $4.11\times10^9$ | $4.10\times10^9$ | $8.35\times10^{-30}$ | $2.64\times10^{-6}$ |
| 28 | 27.08886055 | 14.50815015 | $4.00\times10^{-15}$ | $4.00\times10^{-15}$ | 0.441443826 | $1.14\times10^{-26}$ | $1.14\times10^{-26}$ | 6.72496266 | $4.14\times10^9$ | $4.10\times10^9$ | $1.14\times10^{-26}$ | $6.40\times10^{-8}$ |
| 29 | 32.81824491 | 31.63298352 | $4.00\times10^{-15}$ | $4.00\times10^{-15}$ | 0.474952227 | $6.16\times10^{-34}$ | $6.16\times10^{-34}$ | 9.99826971 | $4.12\times10^9$ | $4.10\times10^9$ | $6.16\times10^{-34}$ | $2.92\times10^{-10}$ |
| 30 | 14.31758698 | 13.11416244 | $4.00\times10^{-15}$ | $4.00\times10^{-15}$ | 0.288049224 | $3.08\times10^{-35}$ | $3.08\times10^{-35}$ | 2.066754282 | $4.13\times10^9$ | $4.10\times10^9$ | $3.08\times10^{-35}$ | $6.01\times10^{-14}$ |

| | F15 | | F16 | | F17 | | F18 | |
|---|---|---|---|---|---|---|---|---|
| | ANA | FDO | ANA | FDO | ANA | FDO | ANA | FDO |
| 1 | 0 | $2.22\times10^{-16}$ | $4.82\times10^{-6}$ | $9.99\times10^{-16}$ | 23.78881433 | 23.68277601 | 223.5535953 | 223.5513726 |
| 2 | $1.15\times10^{-14}$ | 0 | $1.43\times10^{-5}$ | $1.11\times10^{-15}$ | 23.80232686 | 23.93678471 | 223.5667589 | 223.5513726 |
| 3 | $9.99 \times 10^{-16}$ | $9.99 \times 10^{-16}$ | $4.20 \times 10^{-6}$ | $9.99 \times 10^{-16}$ | 23.77803764 | 23.69035899 | 223.5553026 | 223.5513726 |
| 4 | $6.28 \times 10^{-14}$ | $1.11 \times 10^{-16}$ | $6.74 \times 10^{-7}$ | $1.33 \times 10^{-15}$ | 23.74335581 | 23.73681148 | 223.5581154 | 223.5513726 |
| 5 | $1.82 \times 10^{-13}$ | $8.96 \times 10^{-16}$ | $1.34 \times 10^{-6}$ | $1.33 \times 10^{-15}$ | 23.77743535 | 23.7435614 | 223.5626506 | 223.5513726 |
| 6 | $6.22 \times 10^{-15}$ | $1.11 \times 10^{-16}$ | $1.61 \times 10^{-5}$ | $9.99 \times 10^{-16}$ | 23.79436261 | 23.71556709 | 223.5598048 | 223.5513726 |
| 7 | $8.22 \times 10^{-15}$ | $1.11 \times 10^{-16}$ | $1.03 \times 10^{-6}$ | $9.99 \times 10^{-16}$ | 23.71566546 | 23.68432366 | 223.5578919 | 223.5513726 |
| 8 | $1.44 \times 10^{-13}$ | $5.55 \times 10^{-16}$ | $3.48 \times 10^{-6}$ | $8.88 \times 10^{-16}$ | 23.87966915 | 23.70108012 | 223.5612475 | 223.5513726 |
| 9 | $1.11 \times 10^{-16}$ | $1.11 \times 10^{-16}$ | $1.69 \times 10^{-6}$ | $7.77 \times 10^{-16}$ | 23.70607016 | 23.93939471 | 223.5542233 | 223.5513726 |
| 10 | $4.44 \times 10^{-16}$ | $1.33 \times 10^{-15}$ | $6.06 \times 10^{-6}$ | $1.22 \times 10^{-15}$ | 23.71951226 | 23.83709351 | 223.5604607 | 223.5513726 |
| 11 | $1.22 \times 10^{-15}$ | $9.99 \times 10^{-16}$ | $3.76 \times 10^{-6}$ | $7.77 \times 10^{-16}$ | 23.7136252 | 23.93896182 | 223.5635372 | 223.5513726 |
| 12 | $3.75 \times 10^{-14}$ | $1.11 \times 10^{-16}$ | $5.76 \times 10^{-6}$ | $6.66 \times 10^{-16}$ | 23.89952076 | 23.69320027 | 223.5517005 | 223.5513726 |
| 13 | $1.26 \times 10^{-13}$ | $0$ | $9.02 \times 10^{-6}$ | $5.55 \times 10^{-16}$ | 23.83970207 | 23.94535677 | 223.5764026 | 223.5513726 |
| 14 | $3.00 \times 10^{-15}$ | $3.33 \times 10^{-16}$ | $1.58 \times 10^{-6}$ | $7.77 \times 10^{-16}$ | 23.79218864 | 23.76833148 | 223.5804875 | 223.5513726 |
| 15 | $3.11 \times 10^{-15}$ | $4.44 \times 10^{-16}$ | $1.51 \times 10^{-6}$ | $1.22 \times 10^{-15}$ | 23.74107639 | 23.68135525 | 223.5578101 | 223.5513726 |
| 16 | $3.30 \times 10^{-14}$ | $3.33 \times 10^{-15}$ | $8.91 \times 10^{-6}$ | $1.33 \times 10^{-15}$ | 23.75687297 | 23.6900798 | 223.5595445 | 223.5513726 |
| 17 | $1.19 \times 10^{-13}$ | $0$ | $1.04 \times 10^{-5}$ | $1.11 \times 10^{-15}$ | 23.85584942 | 23.81447635 | 223.551743 | 223.5513726 |
| 18 | $1.14 \times 10^{-12}$ | $2.22 \times 10^{-16}$ | $3.24 \times 10^{-6}$ | $5.55 \times 10^{-16}$ | 23.71853807 | 23.73052044 | 223.5573346 | 223.5513726 |
| 19 | $5.55 \times 10^{-15}$ | $0$ | $1.04 \times 10^{-6}$ | $8.88 \times 10^{-16}$ | 23.74365598 | 23.97056993 | 223.5600951 | 223.5513726 |
| 20 | $3.76 \times 10^{-14}$ | $6.66 \times 10^{-16}$ | $5.37 \times 10^{-7}$ | $8.88 \times 10^{-16}$ | 23.79912171 | 23.84108883 | 223.5606669 | 223.5513726 |
| 21 | $1.78 \times 10^{-14}$ | $2.55 \times 10^{-15}$ | $5.21 \times 10^{-6}$ | $9.99 \times 10^{-16}$ | 23.73634589 | 24.43502039 | 223.5672378 | 223.5513726 |
| 22 | $8.88 \times 10^{-16}$ | $1.78 \times 10^{-15}$ | $3.08 \times 10^{-5}$ | $5.55 \times 10^{-16}$ | 23.74978395 | 23.76997022 | 223.5547583 | 223.5513726 |
| 23 | $3.77 \times 10^{-15}$ | $1.11 \times 10^{-16}$ | $3.72 \times 10^{-6}$ | $1.22 \times 10^{-15}$ | 23.81821264 | 23.7822576 | 223.5545055 | 223.5513726 |
| 24 | $1.71 \times 10^{-14}$ | $4.77 \times 10^{-15}$ | $1.88 \times 10^{-6}$ | $1.33 \times 10^{-15}$ | 23.77440552 | 23.75868399 | 223.5644214 | 223.5513726 |
| 25 | $2.22 \times 10^{-16}$ | $2.11 \times 10^{-15}$ | $3.64 \times 10^{-6}$ | $8.88 \times 10^{-16}$ | 23.80311496 | 23.77651152 | 223.5593722 | 223.5513726 |
| 26 | $7.77 \times 10^{-16}$ | $1.11 \times 10^{-16}$ | $1.40 \times 10^{-6}$ | $5.55 \times 10^{-16}$ | 23.7715509 | 24.04397935 | 223.553155 | 223.5513726 |
| 27 | $6.66 \times 10^{-16}$ | $3.33 \times 10^{-16}$ | $5.15 \times 10^{-6}$ | $4.44 \times 10^{-16}$ | 23.72388179 | 23.89049499 | 223.5593623 | 223.5513726 |
| 28 | $9.44 \times 10^{-15}$ | $4.44 \times 10^{-16}$ | $1.29 \times 10^{-6}$ | $7.77 \times 10^{-16}$ | 23.76062509 | 23.7717037 | 223.5567037 | 223.5513726 |
| 29 | $9.21 \times 10^{-15}$ | $2.22 \times 10^{-16}$ | $2.75 \times 10^{-6}$ | $9.99 \times 10^{-16}$ | 23.79965306 | 23.76744563 | 223.5547438 | 223.5513726 |
| 30 | $5.55 \times 10^{-16}$ | $2.22 \times 10^{-16}$ | $9.19 \times 10^{-6}$ | $1.33 \times 10^{-15}$ | 23.84596893 | 23.76024475 | 223.589931 | 223.5513726 |

**Table A6.** ANA and FDO normality test (*p* value) using Shapiro–Wilk test.

| Algorithm | F1 | F2 | F3 | F4 | F5 | F7 | F9 | F10 |
|---|---|---|---|---|---|---|---|---|
| ANA | $8.75 \times 10^{-11}$ | $1.65376 \times 10^{-6}$ | 0.00229666 | $9.48 \times 10^{-12}$ | $5.40 \times 10^{-11}$ | **0.659017** | **0.581344** | $1.10 \times 10^{-11}$ |
| FDO | $9.34 \times 10^{-12}$ | $4.45 \times 10^{-11}$ | $1.09 \times 10^{-10}$ | $2.47 \times 10^{-10}$ | $1.58 \times 10^{-9}$ | 0.0223175 | $4.79981 \times 10^{-5}$ | $4.59 \times 10^{-11}$ |
| | **F11** | **F12** | **F13** | **F14** | **F15** | **F16** | **F17** | **F18** |
| ANA | **0.384774** | $7.40037 \times 10^{-6}$ | $4.32195 \times 10^{-5}$ | $8.86 \times 10^{-12}$ | $7.55 \times 10^{-11}$ | $2.19701 \times 10^{-6}$ | **0.149867** | $6.46068 \times 10^{-5}$ |
| FDO | **0.0515488** | $1.39 \times 10^{-7}$ | NaN | $9.95 \times 10^{-9}$ | $9.53 \times 10^{-12}$ | **0.0560245** | 0.000011896 | $4.27 \times 10^{-13}$ |

**Table A7.** ANA and FDO Homocedasticity test (*p* value) using Levene's test.

| F1 | F2 | F3 | F4 | F5 | F7 | F9 | F10 |
|---|---|---|---|---|---|---|---|
| **$1.33 \times 10^{-1}$** | **0.18407295** | $5.73 \times 10^{-8}$ | **$9.35 \times 10^{-2}$** | **$8.25 \times 10^{-2}$** | **0.89784614** | **0.36480241** | $2.91 \times 10^{-1}$ |
| **F11** | **F12** | **F13** | **F14** | **F15** | **F16** | **F17** | **F18** |
| 0.03998253 | 0.00594079 | $2.7128 \times 10^{-5}$ | $8.29 \times 10^{-3}$ | **$5.58 \times 10^{-1}$** | 0.00033967 | 0.0225408 | $4.3528 \times 10^{-5}$ |